\documentclass[acmsmall]{acmart}
\usepackage{CJKutf8}
\usepackage{multirow}
\usepackage{amsmath,amsfonts}
\usepackage{algorithmic}
\usepackage{graphicx}
\usepackage{booktabs}
\usepackage{subfigure}
\usepackage{textcomp}
\usepackage{xcolor}
\usepackage{color}
\usepackage{comment}
\usepackage{bibentry}
\usepackage{url}
\usepackage{makecell}

\AtBeginDocument{%
  \providecommand\BibTeX{{%
    \normalfont B\kern-0.5em{\scshape i\kern-0.25em b}\kern-0.8em\TeX}}}

\setcopyright{acmcopyright}
\copyrightyear{2020}
\acmYear{2020}
\acmDOI{10.1145/1122445.1122456}

\acmJournal{JACM}
\acmVolume{37}
\acmNumber{4}
\acmArticle{111}
\acmMonth{8}

\begin{document}

\title{Generative Adversarial Networks: A Survey Towards Private and Secure Applications}


\author{Zhipeng Cai}
\email{zcai@gsu.edu}
\author{Zuobin Xiong}
\email{zxiong2@student.gsu.edu}
\author{Honghui Xu}
\email{hxu16@student.gsu.edu}
\author{Peng Wang}
\email{wpeng12@student.gsu.edu}
\author{Wei Li}
\authornote{Corresponding Authors.}
\email{wli28@gsu.edu}
\affiliation{%
	\institution{Georgia State University}
	\city{Atlanta}
	\state{Georgia}
	\country{USA}
	\postcode{30303}
}

\author{Yi Pan}
\email{yipan@gsu.edu}
\authornotemark[1]
\affiliation{%
	\institution{Shenzhen Institute of Advanced Technology, CAS}
	\city{Shenzhen}
	\country{China}
	\postcode{518055}
}
\affiliation{%
	\institution{Georgia State University}
	\city{Atlanta}
	\state{Georgia}
	\country{USA}
	\postcode{30303}
}

\renewcommand{\shortauthors}{Cai, et al.}

\begin{abstract}
Generative Adversarial Networks (GAN) have promoted a variety of applications in computer vision, natural language processing, {\em etc}. due to its generative model's compelling ability to generate realistic examples plausibly drawn from an existing distribution of samples.
GAN not only provides impressive performance on data generation-based tasks but also stimulates fertilization for privacy and security oriented research because of its game theoretic optimization strategy.
Unfortunately, there are no comprehensive surveys on GAN in privacy and security, which motivates this survey paper to summarize those state-of-the-art works systematically.
The existing works are classified into proper categories based on privacy and security functions, and this survey paper conducts a comprehensive analysis of their advantages and drawbacks. 
Considering that GAN in privacy and security is still at a very initial stage and has imposed unique challenges that are yet to be well addressed, this paper also sheds light on some potential privacy and security applications with GAN and elaborates on some future research directions. 
\end{abstract}



\maketitle


\section{Introduction}
\label{sec:intro}
The technological breakthrough brought by Generative Adversarial Networks (GAN) has rapidly produced a revolutionary impact on machine learning and its related fields, and this impact has already flourished to various of research areas and applications.
As a powerful generative framework, GAN has significantly promoted many applications with complex tasks, such as image generation, super-resolution, text data manipulations, {\em etc}.
Most recently, exploiting GAN to work out elegant solutions to severe privacy and security problems becomes increasingly popular in both academia and industry due to its game theoretic optimization strategy.
This survey aims to provide a comprehensive review and an in-depth summary of the state-of-the-art technologies and discuss some promising future research directions for GAN in the area of privacy and security.
We start our survey with a brief introduction to GAN.

\subsection{Generative Adversarial Networks (GAN)}
GAN was first proposed by Goodfellow {\em et al.} to serve as a generative model bridging between supervised learning and unsupervised learning in 2014~\cite{goodfellow2014generative}, which is highly praised as ``the most interesting idea in the last 10 years in Machine Learning'' by Yann LeCun, the winner of 2018 Turing Award. 
Typically, a generative model takes a training dataset drawn from a particular distribution as input and tries to produce an estimated probability distribution to mimic a given real data distribution.
In particular, a zero-sum game between the generator and the discriminator is designed to achieve realistic data generation.
That is, the generator of GAN is trained to fool the discriminator whose goal is to distinguish the real data from the generated data.

GAN has promoted many emerging data-driven applications related to Big Data and Smart Cities thanks to its fantastic properties:
(i) the design of generative models offers an excellent way to capture a high-dimensional probability distribution that is an important research focus in mathematics and engineering domains;
(ii) a well-trained generative model can break through the imprisonment of data shortage for technical innovation and performance improvement in many fields, especially for deep learning; for instance, the high-quality generated data can be incorporated into semi-supervised learning, for which the influence of missing data could be mitigated to some extent;
and (iii) generative models (particularly, GAN) enable learning algorithms to work well with multi-modal outputs, in which more than one correct output may be obtained from a single input for a task, {\em e.g.}, the next frame prediction~\cite{lotter2015unsupervised}.

Prior to GAN, several generative models stemming from the maximum likelihood estimation existed, each of which was state-of-the-art at the time it was proposed.
These prior generative models are either explicit density-based or implicit density-based, depending on whether the underlying distribution can be explicitly pre-defined.
Some well-known explicit density-based models, including Restricted Boltzmann Machine (RBM)~\cite{aliferis2003hiton}, Fully Visible Belief Networks (FVBN)~\cite{frey1996does}, Gaussian Mixture Model (GMM)~\cite{rasmussen2000infinite}, Naive Bayes Model (NBM)~\cite{jiang2008novel}, and Hidden Markov Model (HMM)~\cite{rabiner1989tutorial}, can specialize some specific problems and scenarios but are limited by their common weakness -- 
when the explicitly defined probability density function has intensive parameters and complex dimensions, the computational tractability issue happens where the maximum likelihood estimation may not be able to represent the complexity of the sample data and therefore cannot learn the high-dimension data distribution well.
In addition, the majority of the prior generative models, such as implicit density-based Markov Chain models, require an assumption of Markov Chain that has an ambiguous distribution and can be mixed between patterns.
On the contrary, GAN gets rid of the high-dimension constraint and the Markov Chain dependence.
The generator of GAN uses a pre-defined low-dimension latent code as input and then maps its input to the target data dimension.
Besides, GAN is a nonparametric method and does not require any approximate distribution or Markov Chain property, which endows GAN with the ability to represent the generated data in a lower dimension using fewer parameters. 
Most importantly, GAN is more like an adversarial training framework instead of a rigorous formulation. Thus it is more flexible and extensible to be transformed into many variants according to different requirements, {\em e.g.,} WGAN~\cite{arjovsky2017wasserstein}, InfoGAN~\cite{chen2016infogan}, and CycleGAN~\cite{zhu2017unpaired}.
Motivated by these characteristics, novel research benefits from GAN in a widespread way.

\subsection{The Most Recent Research on GAN}
Currently, two mainstream kinds of research on GAN are being conducted concurrently: application-oriented study and theory-oriented study.
As the restrictions of previous generative models are overcome by GAN, the applications related to data generation are thoroughly investigated for different data formats, such as image generation~\cite{ma2017pose, vondrick2016generating, yang2017high,odena2017conditional}, natural language processing~\cite{fedus2018maskgan,yang2017semi,dai2017good,jetchev2016texture}, time series data generation~\cite{donahue2018synthesizing,hartmann2018eeg,esteban2017real,brophy2019quick}, semantic segmentation~\cite{zhu2016adversarial,luc2016semantic,qiu2017deep,souly2017semi}, {\em etc}.
These application scenarios can be further divided into fine-grained subcategories, including image-to-image translation, image super-resolution, image in-painting, face aging, human pose synthesis, object detection, sketch synthesis, text synthesis, medical data generation, texture synthesis, language and speech synthesis, video and music generation, and so on.
Those prosperous applications demonstrate the extraordinary capability and widespread popularity of GAN.

Meanwhile, to push GAN's capability to a higher level, theoretical methodologies proceed to tackle essential issues including non-stable training, mode collapse, gradient vanish, lack of proper evaluation metrics, {\em etc}.
Some feasible solutions have been proposed, {\em e.g.,} feature matching, unrolled GAN, mini-batch discrimination, self-attention GAN, label smoothing, proper optimizer, gradient penalty, and alternative loss function~\cite{mao2017least}.
%
\begin{figure*}[htb]
	\centering
	\includegraphics[width=\linewidth,height=7cm]{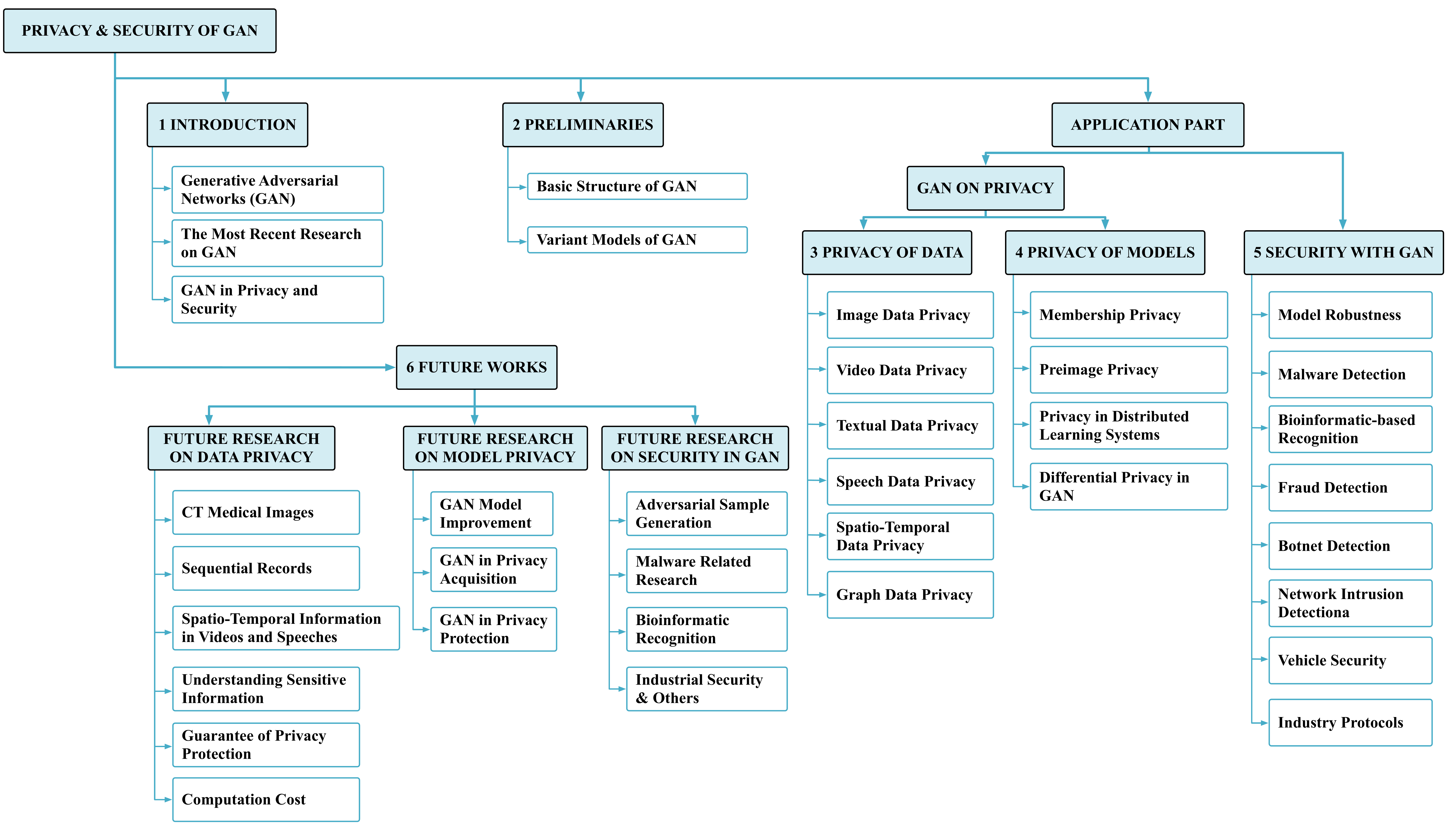}
	\caption{The Organization of This Survey and Its Taxonomy.}
	\label{fig:structure1}
\end{figure*}
%

\subsection{GAN in Privacy and Security}

With individuals' increasing privacy concerns and governments' gradually strengthening privacy regulations, thwarting security and privacy threats has been put in a critical place when designing applications, such as medical image analysis, street-view image sharing, and face recognition.

Thanks to the characteristics of adversarial training, GAN and its variants can be exploited to investigate the privacy and security issues without any pre-determined assumptions of opponents' capabilities that are often hard to be determined in traditional attacks and defense mechanisms.
As the adversarial training process can capture the interactions between an attacker and a defender in a min-max game, the GAN-based methods can be formulated to either launch an attack to break a solid defense or implement protection to defend against strong attackers.
For an attack model, the generator is modeled as an attacker aiming at fooling a defender ({\em i.e.,} the discriminator)~\cite{hitaj2017deep,baluja2017adversarial,zhao2017generating,gao2020improved}. In a defense model, the generator is modeled as a defender to resist a powerful attacker ({\em i.e.}, the discriminator), such as Generative Adversarial Privacy (GAP)~\cite{huang2017context}, Privacy Preserving Adversarial Networks (PPAN)~\cite{tripathy2019privacy}, Compressive Adversarial Privacy (CAP)~\cite{chen2018understanding}, and Reconstructive Adversarial Network (RAN)~\cite{liu2019better}.

In a nutshell, the existing GAN-based privacy and security methods mainly differ in their configurations of GAN models and formulations of loss functions.
However, from the perspective of application scenarios, model design, and data utilization, there is plenty of room for taking maximum advantage of GAN, leaving lots of research blanks for further enhancements.
Those potential directions are elaborated at the end of this survey.

The organization of this survey is illustrated in Fig.~\ref{fig:structure1}.
We present the preliminaries about GAN and its variants in Section~\ref{sec:preliminaries}.
The applications of GAN for privacy, including data privacy and model privacy, are reviewed in Section~\ref{sec:data_privacy} and Section~\ref{sec:model_privacy}, respectively.
The security related applications are described in Section~\ref{sec:security}, and the promising future works are discussed in Section~\ref{sec:future_work}.
Finally, this survey is concluded in Section~\ref{sec:conclusion}.

\section{Preliminaries}
\label{sec:preliminaries}
In this section, we review the basic structure of GAN and its variant models.

\subsection{Basic Structure of GAN}
\label{subsec:Basic_Struct_GAN}

\begin{figure}[h]
	\centering
	\subfigure[The Architecture of Basic GAN]{
		\begin{minipage}[t]{0.5\linewidth}
			\centering
			\label{fig:GAN}
			\includegraphics[width=\textwidth]{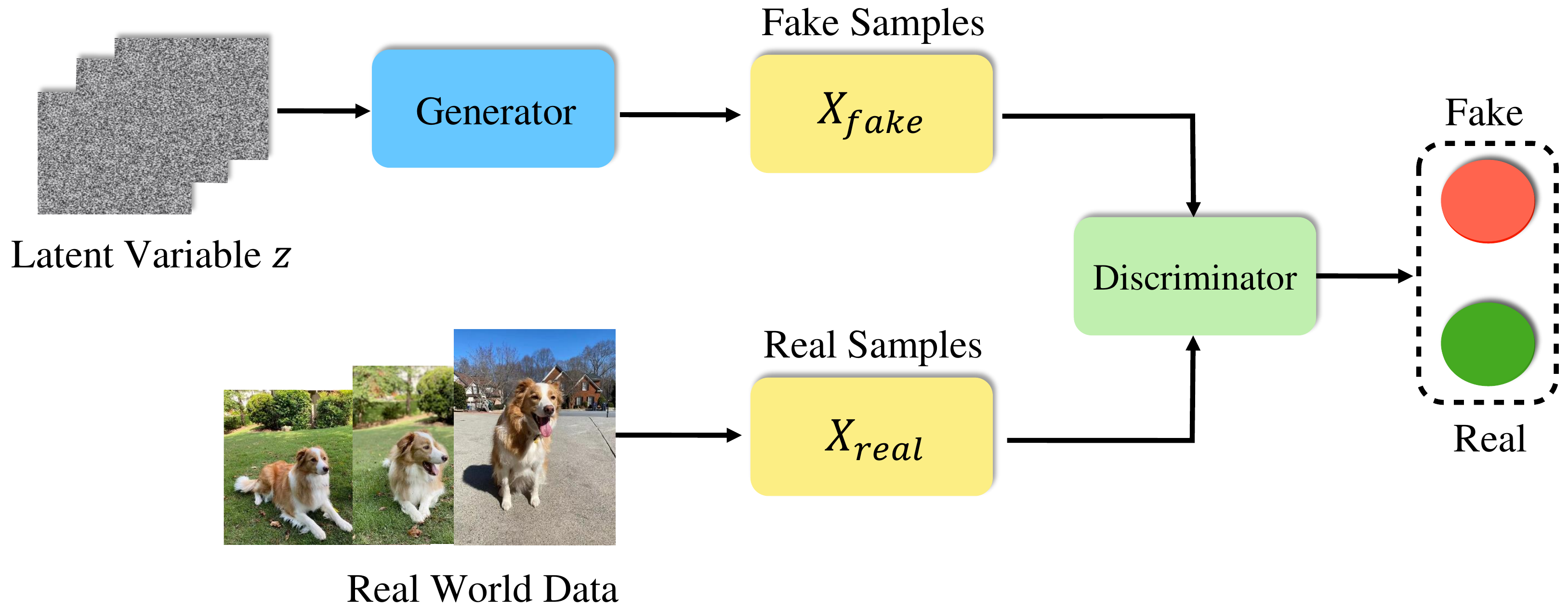}
		\end{minipage}%
	}%
	\subfigure[The Evolution of GAN Models]{
		\begin{minipage}[t]{0.48\linewidth}
			\centering
			\label{fig:Develop_GAN}
			\includegraphics[width=\textwidth]{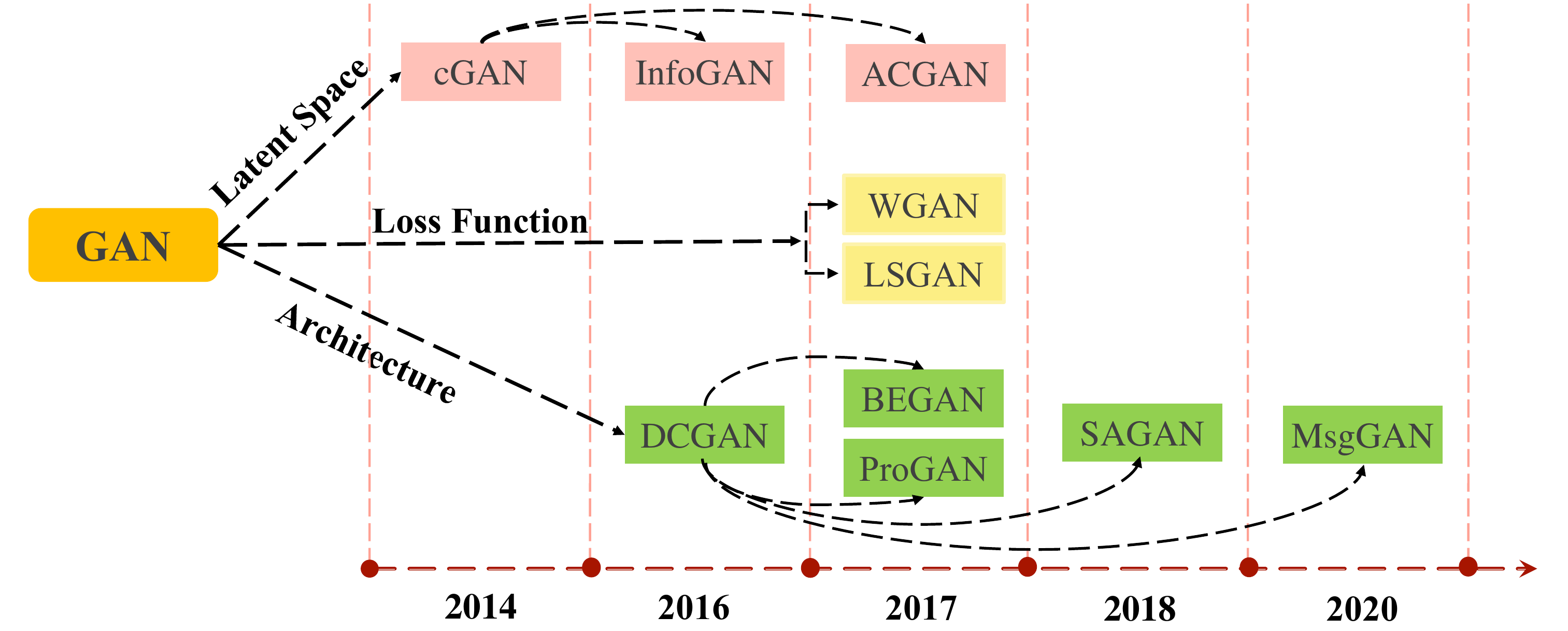}
		\end{minipage}%
	}%
	\caption{The Architecture of GAN and Its Variants.}
\end{figure}

The basic idea of GAN was first proposed in~\cite{goodfellow2014generative}, where a generator can be well trained under an adversarial training framework.
As shown in Fig.~\ref{fig:GAN}, GAN consists of a generator $G$ and a discriminator $D$.
$G$ is a function of operating a latent space $z$ to generate real-like fake data $X_{fake}$, while $D$ is a function to distinguish $X_{fake}$ and real data $X_{real}$.
The training process of $G$ is terminated until $X_{fake}$ and $X_{real}$ are indistinguishable by $D$~\cite{hong2019generative}. 
The interactions between $G$ and $D$ in the adversarial training scenario can be modeled as a min-max game with the following objective:
\begin{equation}
	\label{eq:GAN}
	\min \limits_{G} \max \limits_{D}\mathbb{E}_{x\sim p_{data}}[\log D(x)] + \mathbb{E}_{z\sim p_{z}}[\log (1-D(G(z)))],
\end{equation} 
where $x\sim p_{data}$ denotes the distribution of $X_{real}$, and $z\sim p_{z}$ denotes the distribution of $z$.
%

\subsection{Variant Models of GAN}
\label{subsec:Variants_GAN}
Inspired by the initial design of GAN, a number of variant models have been proposed for various scenarios. In the following, we introduce several popular ones. 

\subsubsection{Wasserstein GAN (WGAN)}
\label{subsubsec:WGAN}
Wasserstein Generative Adversarial Network (WGAN) was developed to solve the problem of mode collapse in a training process to some extent~\cite{arjovsky2017wasserstein}. 
To generate real-looking data that can fool a discriminator, WGAN is trained for minimizing the Wasserstein distance between the real-like data distribution $p_{g}$ and the real data distribution $p_{data}$.

\subsubsection{Least Squares GAN (LSGAN)}
\label{subsec:LSGAN}
To tackle the issue of gradients vanishing in the training process of GAN, the $a$-$b$ coding scheme was utilized in the least squares method to formulate the loss function of discriminator in Least Squares Generative Adversarial Network (LSGAN)~\cite{mao2017least}.
Accordingly, the objective functions of the discriminator and the generator are expressed below respectively:
	\begin{equation}
		\label{eq:LSGAN1}
		\min \limits_{D} \frac{1}{2}\mathbb{E}_{x\sim p_{data}}[(D(x) - b)^2] + \frac{1}{2}\mathbb{E}_{z\sim p_{z}}[(D(G(x)) - a)^2],
	\end{equation} 
	\begin{equation}
		\label{eq:LSGAN2}
		\min \limits_{G} \frac{1}{2}\mathbb{E}_{z\sim p_{z}}[(D(G(x)) - \delta)^2],
	\end{equation} where $\delta$ represents the value that $G$ wants $D$ to classify on fake data.

\subsubsection{Conditional GAN (cGAN)}
\label{subsubsec:cGAN}
Considering that auxiliary information can also assist in generating data, it is natural to extend GAN to a conditional version named Conditional Generative Adversarial Network (cGAN) that provides both the generator and the discriminator with auxiliary information~\cite{mirza2014conditional, miyato2018cgans}.
In cGAN, the latent space $z$ and the auxiliary information $y$ ({\em e.g.}, class labels and data from other modalities) are combined as the conditional input of the generator to make the conditional fake data as similar as the conditional real data.
Accordingly, the objective function of cGAN can be expressed by Eq.~\eqref{eq:cGAN}. 
\begin{equation}
	\label{eq:cGAN}
	\min \limits_{G} \max \limits_{D}\mathbb{E}_{x\sim p_{data}}[\log D(x|y)] + \mathbb{E}_{z\sim p_{z}}[\log (1-D(G(z|y)))].
\end{equation}

\subsubsection{Information Maximizing GAN (InfoGAN)}
\label{subsec:InfoGAN}
Information Maximizing Generative Adversarial Network (InfoGAN) attempts to learn representations with the idea of maximizing the mutual information between labels and the generative data~\cite{chen2016infogan}.
To this end, InfoGAN introduces another classifier $Q$ to predict $y$ given by $G(z|y)$ based on cGAN.
Thus, the objective function of InfoGAN is a regularization of cGAN's objective function, shown as below:
	\begin{equation}
		\label{eq:InfoGAN}
		\min \limits_{G} \max \limits_{D} V(D,G) - \lambda I(G,Q),
	\end{equation} where $V(D,G)$ is the objective function of cGAN except that the discriminator does not take $y$ as input, $I(\cdot)$ is the mutual information and $\lambda$ is a positive hyperparameter.

\subsubsection{Auxiliary Classifier GAN (ACGAN)}
\label{subsubsec:ACGAN}
Auxiliary Classifier Generative Adversarial Network (ACGAN) is a variant of the basic GAN with an auxiliary classifier~\cite{odena2017conditional}, which attempts to learn a representation for $z$ with a class label. 
In ACGAN, every generated sample also has a corresponding class label $c$.
On the other hand, the discriminator is trained to classify its input as real or fake. The auxiliary classifier is used to obtain a probability distribution over the class labels.
Hence, there are two loss functions for training ACGAN: the $\log$-likelihood of the correct source shown in Eq.~\eqref{eq:ACGAN1} and the $\log$-likelihood of the correct label presented in Eq.~\eqref{eq:ACGAN2}.
\begin{equation}
	\label{eq:ACGAN1}
	L_S = \mathbb{E}[\log P(S = real | X_{real})] + \mathbb{E}[\log P(S = fake | X_{fake})].
\end{equation}
\begin{equation}
	\label{eq:ACGAN2}
	L_C = \mathbb{E}[\log P(C = c  | X_{real})] + \mathbb{E}[\log P(C = c | X_{fake})].
\end{equation} 
In Eq.~\eqref{eq:ACGAN1} and Eq.~\eqref{eq:ACGAN2}, $P(S = real | X_{real})$ is the probability of determining the data sample to be real when it is real, and
$P(C = c  | X_{real})$ is the probability of determining the correct class when the data sample is real.
In ACGAN, the generator is trained via maximizing $L_C - L_S$, and the discriminator is trained via maximizing $L_C + L_S$.

\subsubsection{Deep Convolutional GAN (DCGAN)}
\label{subsubsec:DCGAN}
With the success of deep learning models, especially convolutional neural network (CNN)~\cite{krizhevsky2012imagenet}, Deep Convolutional Generative Adversarial Network (DCGAN) has been proposed to generate images and videos efficiently by setting both the generator and the discriminator as CNNs.
DCGAN can even produce higher visual quality images with the help of a CNN-based generator and discriminator~\cite{suarez2017infrared}.

\subsubsection{Boundary Equilibrium GAN (BEGAN)}
\label{subsubsec:BEGAN}
By configuring the discriminator as an autoencoder, Boundary Equilibrium Generative Adversarial Network (BEGAN) was developed in~\cite{berthelot2017began}.
To prevent the discriminator from beating the generator easily, BEGAN learns the autoencoder loss distributions using a loss derived from the Wasserstein distance instead of learning data distributions directly.

In BEGAN, the generator is trained to minimize the loss of image generation in Eq.~\eqref{eq:BEGAN1}, and the discriminator is trained in Eq.~\eqref{eq:BEGAN2} to minimize the reconstruction loss of the real data and maximize the reconstruction loss of the generated images.
\begin{equation}
    \label{eq:BEGAN1}
    L_{G}=L(G(z)).
\end{equation}
\begin{equation}
    \label{eq:BEGAN2}
    L_{D}=L(D(x)) - k_{t} L (D(G(z))).
\end{equation}
In Eq.~\eqref{eq:BEGAN2}, $ k_{t}=k_{t-1}+\lambda_{k}(\beta L(D(x))-L(D(G(z)))$ is a variable that controls the weight of $L(D(G(z)))$ in $L_{D}$, where $\lambda_{k}$ is the learning rate at the $k$-th iteration in the training process, and $\beta = \frac{\mathbb{E}[L(D(G(z)))]}{\mathbb{E}[L(D(x))]}$ balances the efforts allocated to the generator and the discriminator.

	\subsubsection{Progressive-Growing GAN (ProGAN)}
	\label{subsec:ProGAN}
	Progressive-Growing Generative Adversarial Network (ProGAN)~\cite{karras2017progressive} is built based on DCGAN, where both $G$ and $D$ start training with low-resolution images. It gradually increases the model depth by adding new layers to $G$ and $D$ during the training process and ends with the generation of high-resolution images.
	
	\subsubsection{Self-Attention GAN (SAGAN)}
	\label{subsec:SAGAN}
	Traditional CNNs only focus on local spatial information due to the limited receptive field of CNNs, making it difficult for CNN-based GANs to learn multi-class image datasets.
	Self-attention Generative Adversarial Network (SAGAN)~\cite{zhang2019self} is derived from DCGAN to ensure a large receptive field in $G$ and $D$ via a self-attention mechanism so that SAGAN can be used to learn global long-range dependencies for generating multi-class images better.
	
	\subsubsection{Multi-scale gradients GAN (MsgGAN)}
	\label{subsec:MsgGAN}
	When there is not enough overlap in the supports of the real and fake data distributions, gradients passing from $D$ to $G$ become uninformative, making it difficult to exploit different datasets using GAN models.
	Multi-scale gradient Generative Adversarial Network (MsgGAN)~\cite{karnewar2020msg} overcomes this problem by connecting latent space of $G$ and $D$ while training GAN on multiple datasets, in which more information is shared between $G$ and $D$ to make MsgGAN applicable to different datasets.

	{\bf Summary.}
	{The variants mentioned above of GAN can be categorized into three categories based on their improvement focuses, whose evolution is presented in Fig.~\ref{fig:Develop_GAN}.
	(i) {\bf Latent Space.} 
	In cGAN, the labels can be used in the latent space as a kind of extra information for better generation and discrimination of labeled data.
	Based on cGAN, InfoGAN expects to learn representation by maximizing the mutual information between labels and the generative data, while ACGAN tries to learn representation with labels by using an auxiliary classifier.
    (ii) {\bf Loss Function.}
	WGAN uses Wasserstein distance to calculate the loss to solve the problem of mode collapse in GAN, while LSGAN applies the $a$-$b$ coding scheme in the least squares method to the design of $D$'s loss to solve the problem of gradient vanish in GAN.
	(iii) {\bf Architecture.}	
	The generator in DCGAN is a deep CNN-based architecture to generate more real-like images and videos with high visual quality.
	The discriminator in BEGAN is an autoencoder-based architecture to prevent the discriminator from easily beating the generator at the early training stage for fair adversarial training.
	Based on DCGAN, ProGAN gradually increases the depth of $G$ and $D$ in the training process of GAN to generate high-resolution images, SAGAN relies on a self-attention mechanism to obtain global long-range dependency to generate multi-class images, and MsgGAN connects the latent space of $G$ and $D$ while training on multiple datasets to ensure it can be exploited to different datasets.
	
\section{Privacy of Data}
\label{sec:data_privacy}
According to data type, the mainstream applications of GAN in data privacy protection can be classified into six major categories, including image data privacy, video data privacy, textual data privacy, speech data privacy, spatio-temporal data privacy, and graph data privacy, for which a more detailed classification is presented in Fig.~\ref{tab:survey}.
Technically speaking, on the one hand, the generator is designed as a perturbation function to hide the private information and/or trained by one or more discriminators for privacy-preserving data generation. On the other hand, the discriminator is employed to ensure data similarity so that the generated privacy-preserving data is still usable in real applications but is hard to be distinguished from the real data by attackers.

\subsection{Image Data Privacy}
\label{subsec:Image_Data_Privacy}

	As the most popular images used in deep learning, face images contain various individuals' sensitive information, easily causing privacy leakage and thus have received lots of research attentions~\cite{chen2018vgan,deb2019advfaces,yang2020towards, mirjalili2018semi,wu2019privacy,brkic2017face,brkic2017know, cao20193d}.
	Besides, the privacy of medical images~\cite{kim2019privacy} and street-view images~\cite{uittenbogaard2019privacy,xiong2019privacy} have caught research interests in the recent years.
	Furthermore, a number of GAN-based schemes have been developed for image steganography~\cite{meng2019steganography,shu2020encrypted}, image anonymization~\cite{kim2019latent,tang2017automatic,lee2020privacy}, and image encoding~\cite{pittaluga2019learning,yang2018learning,ding2020privacy,nguyen2020autogan}, which indeed can be exploited on any kind of images besides face/medical/street-view images.
	Currently, the study of face images and medical images focuses on a single object, such as one face and one human organ, while the study of street-view images deals with multiple objects, including pedestrians, vehicles, buildings, and so on.
	In the following, the existing works on face images, medical images, street-view images, image steganography, image anonymization, and image encoding are introduced in order.

\begin{figure*}[tb]
	\centering
	\includegraphics[width=0.618\linewidth,height=4.5cm]{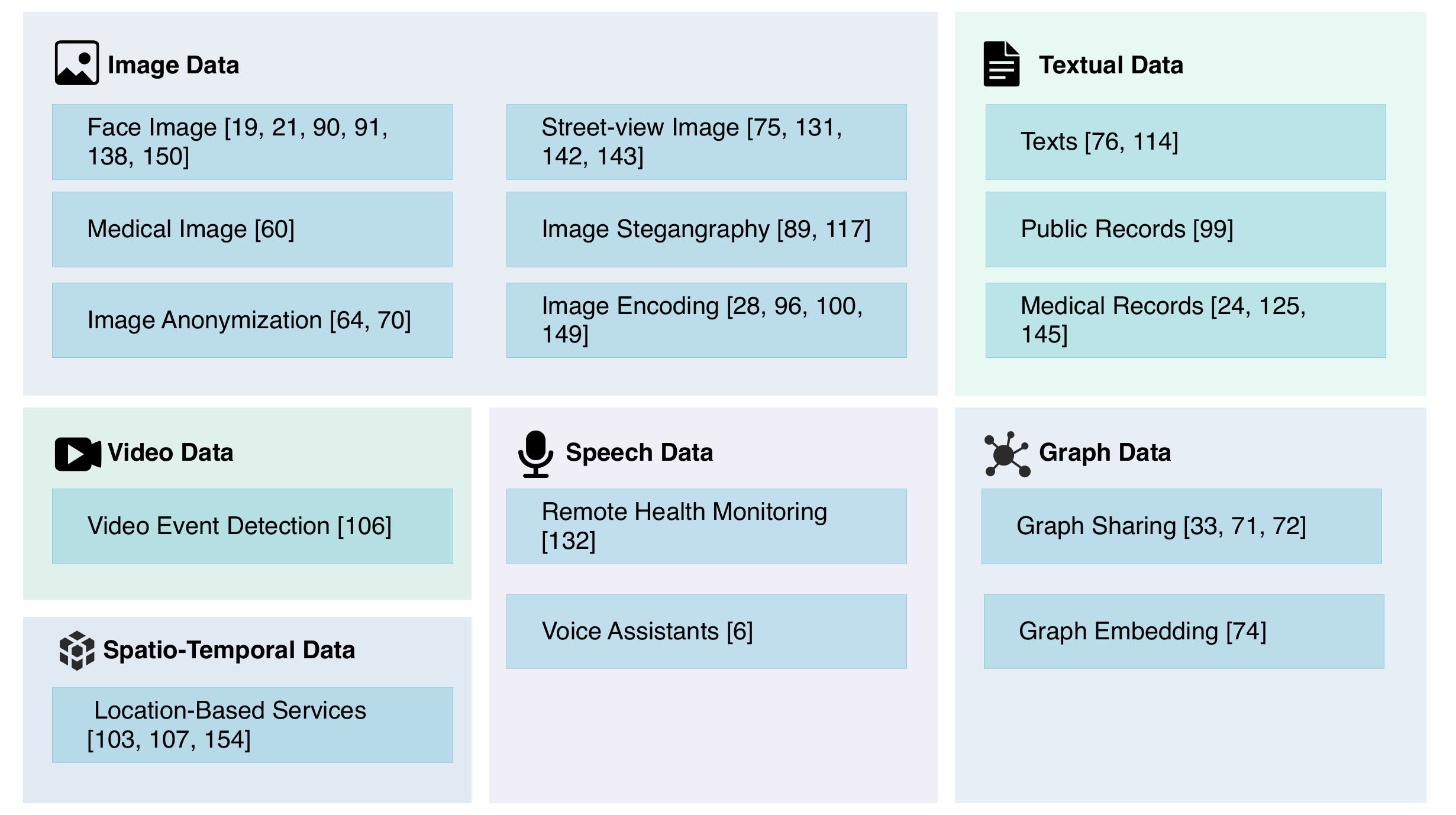}
	\caption{The Applications of GAN-based Data Privacy Protection.}
	\label{tab:survey}
\end{figure*}

	\subsubsection{Face Images}
	Chen {\em et al.}~\cite{chen2018vgan} proposed a method of image representation learning based on Variational Generative Adversarial Network (VGAN) for privacy-preserving facial expression recognition, where Variational Autoencoder (VAE)~\cite{louizos2015variational} and cGAN~\cite{mirza2014conditional} are combined to create an identity-preserving representation of facial images while generating an expression-preserving realistic vision.
	In VGAN, the generator ({\em i.e.,} the encoder-decoder pair in VAE) takes a real image $I$ and its class label $c$ (that indicates a user's identity) as inputs to synthesize a face image.
	Three discriminators in this VGAN model are designed with different functionalities:
	(i) $D_1$ is used for image quality, {\em i.e.}, the synthesized face images should be similar to the real ones;
	(ii) $D_2$ is employed to identity recognition, {\em i.e.}, the identity of the synthetic image should be determined incorrectly by the person identifier;
	and (iii) $D_3$ is exploited for expression recognition, {\em i.e.}, the facial expression in the synthesized data should be guaranteed.
	During the training process, three parameters are set to control the weights of image quality, identity recognition, and expression recognition so that a balance between privacy and utility can be achieved in the synthesized images.
	In addition, to generate identity-preserving face images, Yang {\em et al.}~\cite{yang2020towards} also developed a targeted identity-protection iterative method (TIP-IM) using GAN to generate adversarial identity masks for face images in order to alleviate the identity leakage of face images without sacrificing the visual quality of these face images.
	
	Multi-view identity-preserving face image synthesis ({\em i.e.,} 3D identity-preserving face image generation) has also been studied by Cao {\em et al.}~\cite{cao20193d}. 
	They proposed a DCGAN-based approach to produce realistic 3D photos while preserving identity of multi-view results, in which a face normalizer and an editor are set as the generators to synthesize the 3D photos, and their corresponding discriminators are used to ensure similarity between the synthesized data and the real data.
	This proposed method was demonstrated to dramatically improve the pose-invariant face recognition and generate multi-view face images while preventing the leakage of individuals' identifications.
	
	As well known, real-world recognition systems depend on high-resolution images, which can be used to infer users' identities and biometric information like age, gender, race and health condition through a soft biometric classifier.
	Mirjalili {\em et al.}~\cite{mirjalili2018semi} proposed a model based on ACGAN~\cite{odena2017conditional} to hide the gender information in images for privacy protection.
	In~\cite{mirjalili2018semi}, Autoencoder (AE)~\cite{louizos2015variational} is used as the generator, which is the state-of-the-art method of image generation. 
	The discriminator consists of a 0-1 classifier making the perturbed images to be real-like face images, an auxiliary gender classifier ensuring that the gender attribute of face images is confounded, and a face matcher mitigating the impact on the performance of other biometric recognition.
	
	Based on cGAN, Wu {\em et al.}~\cite{wu2019privacy} designed a model, called Privacy-Protective-GAN (PP-GAN), to preserve soft-biometric attributes during the generation of realistic face with identification.
	Compared with ACGAN, PP-GAN aims at hiding more soft-biometric attributes instead of gender information.
	Moreover, Mirjalili {\em et al. }~\cite{mirjalili2020privacynet} proposed a multi-attribute face privacy model, PrivacyNet, based on GAN to provide controllable soft-biometric privacy protection.
	PrivacyNet allows us to modify an input face image to obfuscate targeted soft-biometric attributes while maintaining the recognition capability on the generated face images.

\subsubsection{Medical Images}

Nowadays, medical data has been widely applied to medical research but possibly suffers from the leakage of individuals' identifications in medical image analysis.
To solve this issue, an adversarial training framework of identity-obfuscated segmentation has been proposed in~\cite{kim2019privacy}. 
Their novel DCGAN-based architecture contains three entities: 
(i) a deep encoder network used as the generator to remove identity features of medical images with the help of additional noise;
(ii) a 0-1 classifier used as a discriminator to guarantee similarity between the encoded images and the original images;
and (iii) a CNN-based medical image analysis network used as another discriminator to analyze image segmentation content.
	This design integrates an encoder, a 0-1 classifier, and a segmentation analysis network to protect medical data privacy and simultaneously maintain medical image segmentation performance.

\subsubsection{Street-View Images}

Street-view services, such as Google Street View and Bing Maps Streetside, typically serve users through collecting millions of images, which some individuals often refuse due to serious privacy concerns~\cite{larson2018pixel}. 
Uittenbogaard {\em et al.}~\cite{uittenbogaard2019privacy} designed a multi-view GAN model based on DCGAN, where the generator is used to detect, remove and paint in moving objects by using multi-view imagery, and the discriminator is used to make the generated images photorealistic.
With these settings, the multi-view GAN removes private regions and is able to retain the utility of the synthesized street-view images.
Similarly, Li {\em et al.} proposed Picprivacy model~\cite{li2020generative} to segment and erase sensitive information, such as human portrait, from the street-view images while repairing blank regions based on GAN to maintain the performance of 3D construction.
Besides, in order to defend against location inference attacks on vehicular camera data, Xiong {\em et al.}~\cite{xiong2020adgan,xiong2019privacy} proposed three Auto-Driving GAN (ADGAN) models based on DCGAN to generate privacy-preserving vehicular images and videos for autonomous vehicles.
The core idea of their three methods is to prevent the location-related background information in images/videos from being identified by attackers and maintain data utility simultaneously.
The generator takes original data as input and outputs the privacy-preserving data, and multiple discriminators are constructed following the convolutional 0-1 classifier structure with different filter sizes to distinguish real/fake data more efficiently. 
Additionally, for the tradeoff between privacy and utility, customized privacy loss and utility loss are calculated through the difference between the original data and the generated data.
To improve model performance and data quality, an extra target model was added in~\cite{xiong2020adgan} to provide more accurate feedback on data generation.

\subsubsection{Image Stegangraphy}
\label{subsubsec:Image_Stegangraphy}
With the widespread Internet of Things (IoT) applications in recent years, the risk of privacy leakage has increased.
Traditionally, steganography is a critical method to find the trade-off between personal privacy disclosure and covert communication.
A new steganography algorithm is developed based on image-to-image translation using cyclic DCGAN framework, where $G_{1}$ is a steganography module transferring data from $x_1$-domain to $x_2$-domain, and $G_{2}$ is another steganography module transferring data from $x_2$-domain to $x_1$-domain.
These two steganography modules are used as two generators in Steganography-CycleGAN so that the stego images generated by the proposed method will be close to the cover images.
Two discriminators $D_{x_2}$ and $D_{x_1}$ are used to make sure that not only the stego images from $x_1$-domain to $x_2$-domain but also the stego images from $x_2$-domain to $x_1$-domain are similar to the real ones.
Also, in order to resist detection by steganalysis module, one steganalysis module $D_{S}$ is deployed to realize the concealment and security in transmission by an adversarial training.
Similarly, Shu {\em et al.}~\cite{shu2020encrypted} applied GAN to successfully achieve the encrypted rich-data steganography during transmission in wireless networks.

\subsubsection{Image Anonymization}
Protecting individuals' data privacy is an essential task for public data collection and publication. 
In~\cite{kim2019latent}, a privacy-preserving adversarial protector network (PPAPNet) was designed as a DCGAN-based anonymization method that converts a sensitive image into a high-quality and attack-immune synthetic image.
Under PPAPNet, the generator is initialized as a protector with pre-trained Autoencoder, the discriminator is the WGAN~\cite{gulrajani2017improved} critic with a gradient to guide the protector to generate realistic images that can defend model inversion attacks, and a noise amplifier inside the protector plays a vital role in noise optimization for effective image anonymization.
Similarly, Lee {\em et al.}~\cite{lee2020privacy} implemented face anonymization on the drone patrol systems to hide the sensitive information in face images by converting a sensitive image into another synthetic image based on DCGAN.

\subsubsection{Image Encoding}
A formula was established for learning an encoding function based on DCGAN in~\cite{pittaluga2019learning}. 
The encoder is trained to prevent privacy inference and maintain the utility of predicting non-private attributes. 
In this adversarial framework, the generator is an encoding function that outputs limited information of private attributes while preserving non-private attributes, and the discriminators are neural network-based estimators for privacy protection. 
Similarly, the works of~\cite{yang2018learning, ding2020privacy, nguyen2020autogan} attempted to learn representations or dimension-reduced features from raw data based on DCGAN, in which the desired variables are maintained for utility representations, and the sensitive variables are hidden for privacy protection.
Especially, these representations encoded from users' data can exhibit the predictive ability and protect privacy.

\subsection{Video Data Privacy}
\label{subsec:Video_Data_Privacy}
There is an increasing concern in computer vision devices invading users' privacy by recording unwanted videos~\cite{ryoo2016privacy,wang2019privacy,meng2019steganography,xiong2019privacy}.
On the one hand, some previous works that focus on image privacy can be applied to hide sensitive contents in videos by simply considering a video as a sequence of image frames~\cite{yang2018learning,kim2019latent}.
On the other hand, videos can also be used to recognize important events and assist human's daily lives by advanced deep learning models, which differ from image recognition applications.
Thus, it is expected that in videos, individuals' privacy should not be intruded and the efficiency of detecting continuous action should be kept at the same time.
A video face anonymizer is built based on the DCGAN model by~\cite{ren2018learning}, where the generator is designed as a modifier of the face in videos, and two discriminators, including one 0-1 classifier and one face classifier, are deployed.
Specifically, the 0-1 classifier is applied for adversarial training, and the face classifier is used for face detection. 
As a result, the video face anonymizer can finally hide the faces but maintain the information used for action recognition.

\subsection{Textual Data Privacy}
\label{subsec:Textual_Data_Privacy}
To implement privacy protection using GAN for textual data, various schemes for anonymous text synthesis~\cite{shetty2018a4nt,li-etal-2018-towards} and privacy-preserving public/medical records release~\cite{park2018data, choi2017generating,torfi2020corgan,yale2020generation,lee2020generating} have been proposed.

\subsubsection{Texts}
Natural Language Processing (NLP)~\cite{ibrahim2010class} enables author identification of anonymous texts by analyzing the texts' stylistic properties, which has been already applied to describe users by determining their private attributes like age and/or gender. 
Shetty {\em et al.}~\cite{shetty2018a4nt} proposed an author adversarial attribute anonymous neural translation (A4NT) with a basis of DCGAN to defend NLP-based adversaries. 
The objective of A4NT is to fool the identity classifier by altering the semantics of the input text in person while maintaining semantic consistency. 
To this end, the A4NT network is designed as a style-transfer network that transforms texts into a target style based on LSTM~\cite{sundermeyer2012lstm} and fools the attribute classifier simultaneously. 
The generator in an A4NT network transforms the input texts from a source attribute class in order to generate the style of a different attribute class.
One discriminator uses 0-1 classifier to help the generator hide the author's identity, and the other discriminator makes the text semantic consistent by minimizing the semantic and language loss.
With a similar idea, Li {\em et al.}~\cite{li-etal-2018-towards} presented a method to learn text representations instead of texts, preserve users' personal information, and retain text representation utility.

\subsubsection{Public Records}
When sharing records with partners an/or releasing records to public, traditional approaches perform privacy protection by removing identifiers, altering quasi-identifiers and perturbing values. 
In~\cite{park2018data}, the DCGAN architecture with an auxiliary classifier is exploited to develop a model called AC-DCGAN, where the generator produces synthetic records to hide sensitive information.
Through adversarial training, the auxiliary classifier is used to predict synthetic records' labels such that the records' identification cannot be re-identified, and the discriminator is trained to ensure that the synthetic records have similar distributions of real records. 
Thus, the fake and synthetic records can defend re-identification attacks~\cite{brkic2017face} while achieving high utility.

\subsubsection{Medical Records}
Accessing Electronic Health Record (EHR) records data has promoted computational advances in medical research and raised people's privacy concerns about their EHR data. 
Choi {\em et al.}~\cite{choi2017generating} constructed a medical Generative Adversarial Network (medGAN) based on the basic structure of GAN to generate privacy-preserving synthetic patient records. 
MedGAN can generate high-dimensional discrete variables, in which an autoencoder network is used as the generator to produce the synthetic medical data with the help of additional noise, and a 0-1 classifier is used as the discriminator to ensure data similarity. 
As a result, the synthetic medical data is applicable to distribution statistics, predictive modeling, medical expert review, and other medical applications.
A limited privacy risk in both identity and attributes can be achieved using medGAN.
Moreover, to improve the performance of privacy protection of medGAN, the evaluation of privacy-preserving medical records of medGAN was investigated by~\cite{yale2020generation}. CorGAN was developed by~\cite{torfi2020corgan} taking into account the correlations of medical records, and a dual-autoencoder was configured as the generator in medGAN to generate sequential electronic health records instead of discrete records for higher predictive accuracy to assist medical experts~\cite{lee2020generating}.

\subsection{Speech Data Privacy}
\label{subsec:Speech_Data_Privacy}
The works on privacy-preserving speech data based on GAN mainly focus on two fields: remote health monitoring~\cite{vatanparvar2020adapting} and voice assistants in IoT systems~\cite{aloufi2019emotionless}.

\subsubsection{Remote Health Monitoring}
Remote health monitoring has been introduced as a solution to continuous diagnosis and trace of subjects' condition with less effort, which can be partially achieved by passive audio recording technology that may disclose subjects' privacy.
In~\cite{vatanparvar2020adapting}, Vatanparvar {\em et al.} designed a GAN-based speech obfuscation mechanism for passive audio recording when using remote health monitoring.
In this speech obfuscation model, the generator is employed to map the audio recording into the distribution of human speech audio and filter the private background voice, and the discriminator is one 0-1 classifier to determine the probability of human speech presence within the audio.
After the adversarial training, the synthetic audio recording can be obtained to match the human speech distribution for medical diagnosis and avoid the trace of private information.

\subsubsection{Voice Assistance}
Voice-enabled interactions provide more human-like experiences in many popular IoT systems. 
Currently, many speech recognition techniques are developed to offer speech analysis services by extracting useful information from voice inputs as the voice signal is a rich resource containing various states of speakers, such as emotional states, confidence and stress levels, and physical conditions. 
With the voice signal, service providers can build a very accurate profile for a user through the voice, which on the other hand, may lead to privacy leakage.
In~\cite{aloufi2019emotionless}, a cyclic GAN model was built to translate voice from one domain into another domain to hide the users' emotional states in voice, in which the generators are used to do voice translation, and the discriminators are used to force generators to produce the synthetic voice with high quality. 
The synthetic voice can still be successfully exploited to perform speech recognition for voice-controlled IoT services while resisting inference on users' emotional states.

\subsection{Spatio-Temporal Data Privacy}
\label{subsec:Spatio_Temporal_Data_Privacy}
The popularity of edge computing accelerates the emergence and innovation of IoT applications and services.
Since various spatio-temporal data need to be collected from IoT devices ({\em e.g.}, GPS) and the data contains a lot of users' sensitive information, privacy issues are raised unavoidably~\cite{yin2018gans,rezaei2018protecting}.

In~\cite{yin2018gans}, a GAN-based training framework was designed to protect data privacy in two real world mobile datasets, where the generator is trained to learn the features of data for privacy-preserving sharing, and the discriminator is used to guarantee the utility of the synthesized data.
Considering the limited computation capacity of IoT devices, Rezaei {\em et al.}~\cite{rezaei2018protecting} created a privacy-preserving perturbation method that can efficiently run on IoT devices by combining deep learning network and the basic structure of GAN.
They implemented one generator to add noise and two discriminative classifiers (including a target classifier and a sensitive classifier) for adversarial training. 
More concretely, the target classifier attempts to maintain the utility of the mobile data. The sensitive classifier tries to help hide sensitive information during the data generation process, aiming to find a good trade-off between utility and privacy for mobile data in IoT.
In location-based services, the privacy of spatio-temporal trajectories submitted from IoT devices was studied by Rao {\em et al.}~\cite{rao2020lstm}.
The authors proposed a LTSM-TrajGAN model based on DCGAN to generate privacy-preserving synthetic trajectory data, in which the generator is based on LSTM recurrent neural network trained by minimizing the spatial and temporal similarity loss and the discriminator is a 0-1 classifier for performing the adversarial training. 
LTSM-TrajGAN can produce privacy-preserving synthetic trajectory data to prevent reidentification of users and preserve the essential spatial-temporal characteristics of trajectory data.

\begin{table}
	\caption{Comparison of GAN-based Mechanisms for Data Privacy Protection.}
	\label{tab:Data_Privacy}
	\resizebox{\linewidth}{!}{
		\begin{tabular}{ccccccc}
			\toprule
			\textbf{Literature} & \textbf{Application} & \textbf{Input} & \textbf{Output} & \textbf{Model} & \textbf{Data Utility} & \textbf{Data Privacy} \\
			\midrule
			~\cite{chen2018vgan} & Expression Recognition & Face Images & Synthetic Face Images & VGAN & Expression Recognition & Identity \\
			~\cite{yang2020towards} & Face Image Synthesis & Face Images & Synthetic Face Images & TIP-IM & Face Synthesis & Identity \\
			~\cite{cao20193d} & 3D Face Image Synthesis & Face Images & 3D Synthetic Face Images & AD-GAN & 3D Face Synthesis & Identity \\
			~\cite{mirjalili2018semi} & Face Recognition & Face Images & Synthetic Face Images & ACGAN & Face Recognition & Gender \\
			~\cite{wu2019privacy,mirjalili2020privacynet} & Face Recognition & Face Images & Synthetic Face Images & PP-GAN & Face Recognition & Soft-biometric Attributes \\
			~\cite{kim2019privacy} & Medical Image Analysis & Medical Images & Synthetic Medical Images & DCGAN & Image Segmentation & Identity \\
			~\cite{uittenbogaard2019privacy,li2020generative} & Street Image Synthesis & Street Images & Inpainted Street Images & DCGAN & Street Image Synthesis & Private Regions \\
			~\cite{xiong2020adgan,xiong2019privacy} & Autonomous Vehicles & Camera Data & Perturbed Camera Data & AD-GAN & Camera Data Synthesis & Location \\
			~\cite{meng2019steganography,shu2020encrypted} & Image Steganography & Images & Steganographic Images & cyclic DCGAN & - & - \\
			~\cite{kim2019latent,lee2020privacy} & Image Anonymization & Images & Anonymized Images & DCGAN & - & - \\
			~\cite{pittaluga2019learning,yang2018learning,ding2020privacy,nguyen2020autogan} & Image Encoding & Images & Image Representations & DCGAN & - & - \\
			~\cite{ren2018learning} & Action Detection & Video & Face-anonymized Video & DCGAN & Action Dection & Face \\
			~\cite{shetty2018a4nt} & Text Synthesis & Texts & Synthetic Texts & A4NT & Text Synthesis & Identity \\
			~\cite{li-etal-2018-towards} & Text Representation & Texts & Text Features & LSTM-GAN & Text Representation & Identity \\
			~\cite{park2018data} & Record Release & Public Records & Synthetic Records & AC-DCGAN & Record Synthesis & Identity \\
			~\cite{choi2017generating,yale2020generation,torfi2020corgan} & Medical Record Sharing & EHR Records & Synthetic EHR Records & medGAN & Record Synthesis & Identity \\
			~\cite{vatanparvar2020adapting} & Health Monitoring & Audio & Synthetic Audio & Obfuscation & Audio Synthesis & Background Audio \\
			~\cite{aloufi2019emotionless} & Voice Assistance & Voice Signal & Synthetic Voice & cyclic GAN & Voice Synthesis & Emotional States \\
			~\cite{yin2018gans,rezaei2018protecting} & Data Sharing & Mobile Data & Synthetic Mobile Data & Perturbation & Mobile Data Synthesis & Sensitive Information \\
			~\cite{rao2020lstm} & Location-based Services & Trajectories & Synthetic Trajectories & LTSM-TrajGAN & Trajectories Synthesis & Identity \\
			~\cite{fang2019gdagan,li2020adversarial2,li2020graph} & Graph Sharing & Graph & Anonymized Graph & Perturbation & Graph Synthesis & Communities \\
			~\cite{li2020adversarial1} & Graph Embedding & Graph & Graph Representations & APGE & Representations Synthesis & Private Attributes \\
			\bottomrule
		\end{tabular}
	}
\end{table}

\subsection{Graph Data Privacy}
\label{subsec:Graph_Data_Privacy}
The graph data (such as social networks) promotes the research and applications of data mining, but privacy leakage in graph data is also becoming more serious during data processing and sharing procedures.
Although the traditional anonymization methods for the graph data can balance data utility and data privacy to some extent, these methods are vulnerable to the state-of-the-art inference approaches using graph neural networks~\cite{li2020seed}.
Therefore, more powerful strategies are desired to defend inference attacks for graph data.

\subsubsection{Graph Sharing}
Fang {\em et al.}~\cite{fang2019gdagan} developed a Graph Data Anonymization using Generative Adversarial Network (GDAGAN) that exploits the LSTM-based generator for data generation and the 0-1 classifier based discriminator for utility guarantee.
Besides, the Laplace noise is added into the synthetic graph for perturbation to protect privacy before publishing graph data to the public.
The idea of an adversarial graph has been extended in~\cite{li2020adversarial2} to consider both the problems of imperceptible data generation and community detection for an enhanced privacy protection.
The proposed GAN-based model in~\cite{li2020adversarial2} has three critical components:
(i) a constrained graph generator based on graph neural network to generate an adversarial graph;
(ii) a 0-1 classifier working as the discriminator to make the synthetic graph real-like for maintaining utility; 
and (iii) a community detection model that helps the adversarial graph prevent community detection attacks.
Similarly, the graph feature learning model of~\cite{li2020graph} was designed based on GAN to perturb a probability adjacency matrix with the help of Laplace noise in the graph reconstruction process to obtain an anonymous graph.
This reconstructed anonymous graph maintains the utility of link prediction due to GAN's good feature learning ability, and can be used to defend community detection and de-anonymization attack owing to the utilization of Laplace noise.

\subsubsection{Graph Embedding}
It is well known that graph embedding is useful to learn low-dimension feature representations for various prediction tasks.
Adversaries can also infer sensitive information from these graph node representations, resulting in privacy leakage.
Li {\em et al.}~\cite{li2020adversarial1} designed an Adversarial Privacy Graph Embedding (APGE) training framework based GAN to remove users' private information from the learned representations of graph data.
In APGE, one autoencoder-based generator is used to learn graph node representations while implementing disentangling and purging mechanisms.
During the process of adversarial training, one 0-1 classifier is employed to make the synthetic representations real-like, and one non-private attribute prediction model and one private attribute prediction model are designed to keep data utility and protect users' privacy, respectively.

Finally, the comparison of surveyed approaches for data privacy is summarized in Table~\ref{tab:Data_Privacy}.


\section{Privacy of Models}
\label{sec:model_privacy}

In the previous section, we discuss the works on the privacy issues of various sensitive data. 
It is worth noticing that privacy can be inferred not only through data but also through the adopted models, especially in Machine Learning as a Service (MLaaS)~\cite{ribeiro2015mlaas}. 
As analyzed in~\cite{fredrikson2014privacy}, a model's privacy breaches if an adversary can use the model's output to infer the private attributes that is used to train the model.
This section will survey how to steal privacy from the learning models and how to protect the learning model privacy using GAN-based approaches.

\subsection{Membership Privacy}
\label{sec:member:privacy}

Membership inference attacks can be launched towards a machine learning and/or deep learning model to determine if a specific data point is in the given model's training dataset or not~\cite{shokri2017membership}. 
Typically, after a model is trained, an attacker feeds data into the model and gets the corresponding prediction results that can be used as additional knowledge to perform black-box membership inference attacks. 
Such an attack will cause privacy leakage and even other severe consequences. 
For instance, with a patient's medical records and a predictive model trained for a disease, an attacker can know whether the patient has a certain disease by implementing membership inference attacks.
To defend against such attacks, the techniques of GAN, anonymization, and obfuscation have been exploited to design countermeasures~\cite{dwork2006calibrating, al2019privacy}.

\subsubsection{Attacks on Membership Privacy}
Membership privacy of generative models was studied for the first time in ~\cite{hayes2019logan}, in which a model named ``LOGAN'' was designed to attack a generative model through either black-box or white-box via released API in MLaaS. 
In white-box attacks, an attacker is assumed to know the target GAN model's structure and parameter consisting of a generator and a discriminator.
It is known that the discriminator is able to assign a higher probability to a data point that is in the training dataset.
Accordingly, an attacker inputs several data points into the discriminator obtaining the corresponding probabilities and selects the $n$ most probable data points as the $n$ members of the training dataset.
For black-box attacks, membership privacy can be inferred without or with auxiliary knowledge.
If there is no auxiliary knowledge, an attacker uses the generator of the target GAN from query API to produce enough generated data labeled as ``real'' for training a local GAN such that he/she can get a parameterized discriminator.
Then, the attacker performs the aforementioned white-box attacks on his/her discriminator to learn membership privacy.
If the attacker knows some auxiliary information, such as the data only from the training dataset or the data from both the training and test datasets, the attacker's discriminator can be better trained and used in while-box attacks for membership inference.
However, there are too many assumptions in black-box attacks, which may be impractical in real applications.

In~\cite{Liu2018PerformingCA}, the authors proposed ``co-membership'' attack towards generative models. 
Unlike the previous works that infer the membership of a single data point each time, the co-membership attack aims to simultaneously decide whether $n$ ($n\geq 1$) data points are in the training dataset of the target generative model.
The implementation of co-membership attacks comes from the intuitive understanding of GAN: if GAN is powerful enough, it should be able to generate any data from a latent vector or reconstruct any data from its latent representation. 
To accomplish such attacks, a neural network is trained on the attacker side with the objective:
$\min_{\gamma} \frac{1}{n}\sum_{i}^{n} \Delta (x_i,G(A_{\gamma}(x_i)))$,
where $\gamma$ is a network parameter, $G$ is the generator of GAN that takes a latent vector $z$ as input and outputs generated data $G(z)$, and $A_{\gamma}$ is the attacker's network that takes original data $x_i$ as input and outputs a low dimensional vector with a shape same as $z$.
After the training process is finished, the attacker gets a distance ({\em e.g.,} $L_2$ distance) $\Delta (x_i,G(A_{\gamma}(x_i)))$ between $x_i$ and $G(A_{\gamma}(x_i))$. 
If the distance is greater than a threshold, the reconstruction of $x_i$ from $z$ is not associated with the original $x_i$, which means $x_i$ may not be a member of the training dataset; otherwise, $x_i$ is a member of the training dataset.
The proposed attack method has some fatal flaws: (i) for a large training dataset, if the pre-determined value of $n$ is small, the attack accuracy is lower than that of the traditional membership attacks;
and (ii) for each victim model, an attacker needs to train a different attack network $A_{\gamma}$ from randomly initialized weights. 
So the attack efficiency is not as high as expected. 
Besides, the proposed co-membership attack is a kind of white-box attack and is impossible to be used in a black-box scenario.

\subsubsection{Protection of Membership Privacy}
\label{sec:protect:membership}
To prevent membership inference attacks, Nasr {\em et al.}~\cite{nasr2018machine} proposed an end-to-end method that trains a machine learning model with membership privacy protection using adversarial regularization based on GAN, which enables a user to train a privacy-preserving predictive model on the MLaaS platforms ({\em e.g.,} Google, Amazon, and Microsoft). 
Their proposed method contains two parties: an attacker $h$ and a defensive classifier $f$. 
The attacker's privacy gain from the victim model $f$ is defined as:
\begin{equation}
\label{eq:member:attack}
		G_f(h)=\mathop{\mathbb{E}}\limits_{(x,y)\in D(X,Y)}[\log(h(x,y, f(x)))]+\mathop{\mathbb{E}}\limits_{(x,y)\in D'(X,Y)}[\log(1-h(x,y, f(x)))],
\end{equation}
where $(x,y)$ is a data point, $D(X,Y)$ is a training dataset, and $D'(X,Y)$ is the set of data that is not in $D(X,Y)$. 
By maximizing Eq.~\eqref{eq:member:attack}, an attacker can obtain an accurate prediction on all the data points and know if they are in the training dataset.
A defender's loss function is formulated as:
$\min_f(L_D(f)+\lambda \max_h G_f(h))$,
where $L_D$ is the normal loss when training a classifier ({\em e.g.}, cross-entropy), and $\lambda$ is a parameter to adjust the tradeoff between utility and privacy.
In this method, the defensive classifier has two objectives: (i) minimizing the normal loss function of $f$; and (ii) reducing the inference gain $G_f(h)$.
Moreover, $G_f(h)$ also works as a regularization term controlled by $\lambda$, improving the generalization capability of classifier $f$.
The min-max optimization can train a private classifier even if the attacker has the strongest inference gain.
Nevertheless, the proposed method has its disadvantages. 
As shown in the experiments, the trained classifier's classification accuracy is decreased by around $3\%$ compared with the classifiers without an adversarial regularization term. 
Another flaw is that the training process of the proposed method requires much data as a reference dataset. 
In practice, it is hard to obtain so much data with the same distribution as the data given by users in MLaaS, which lowers the applicability of the proposed method.

In~\cite{wu2019generalization}, Wu {\em et al.} investigated the generalization ability of GAN from a novel perspective of privacy protection.
They theoretically analyzed the connection between the generalization gap and the membership privacy for a series of GAN models. 
Motivated by a well-known intuition~\cite{yeom2018privacy}: ``the smaller the generalization gap is, the less information of the training dataset will be revealed'', they linked the stability-based theory and differential privacy~\cite{dwork2006calibrating}, which illustrates that a differentially private training mechanism can not only reduce the membership privacy leakage but also improve the generalization capability of the model.

\subsection{Preimage Privacy}
\label{sec:preimage_privacy}
Some other attacks ({\em e.g.,} model inversion attacks~\cite{fredrikson2015model} and data reconstruction attacks~\cite{feng2010fingerprint}) move one more step and can cause more serious damage to machine learning models.
In model inversion attacks, given a target model $f$ and a label $y_t$, the purpose of an attacker is to retrieve the input $x$ of the target model $f$ such that $f(x)=y_t$.
Similarly, in data reconstruction attacks, an attacker focuses on recovering the raw data in the training dataset of a given model $f$ with the help of additional information.
The objective of these two types of attacks is to find private information of input data of learning models, for which we propose a new term called ``{\em Preimage Privacy}'' to depict model inversion attacks and data reconstruction attacks.

\subsubsection{Attacks on Preimage Privacy}

Model inversion attacks have been successfully conducted as a severe threat under the white-box setting. 
While for the black-box scenario, there are no impressive works before the birth of GAN. 
In~\cite{Avodji2019GAMINAA}, a model inversion attack framework was built under the black-box setting. 
Given a target model $f$ and a label $y_t$, an attacker aims at characterizing data $x_t$ belonging to $y_t$. 
To achieve the goal, an attacker trains a GAN framework using adaptive loss in BEGAN~\cite{berthelot2017began}, where the generator $G$ works as a data inverter, {\em i.e.}, $G: z \to x $, and the discriminator $D$ is replaced by a neural network classifier taking $x$ as input and predicting a label $y$ as output.
Since there is no real data, a randomly sampled dataset $X_D \sim \mathcal{N}(0,1)$ is used for self-adaptive updating based on BEGAN.
The training process of an attack is expressed below:
\begin{equation}
	\label{eq:began_1}
	\min_{G}H(f(G(z)), y_t).
\end{equation}
\begin{equation}
	\label{eq:began_2}
	\min_{D}H(D(X_D), f(X_D))-k_t H(D(G(z)), f(G(z))).
\end{equation}

In Eq.~\eqref{eq:began_1} and Eq.~\eqref{eq:began_2}, $H$ denotes the cross-entropy loss, and $k_t$ is a parameter of self-adaptive updating.
This attack method can efficiently attack the black-box model even though the model is trained with a differential privacy mechanism, providing much inspiration to future research. 
But this method has an essential flaw that is common for all the black-box attacks: an attacker has to issue lots of queries to get predicted label of input data as auxiliary training information, bringing a huge cost.
For example, in this work, 1,280,000 queries are needed to achieve the desired attack performance. 
Such a frequent and intensive query operation may be detected by the target model easily.
Thus, there should be some other ways to improve the attack method.

Moreover, when a target model ({\em e.g.,} a neural network) has high-dimension input, it is difficult to obtain an optimal solution only with the given label $y$ for the attack model, making the resulted $x$ like a random noise in high-dimension space. 
Thus, the results of GAN-based model inversion attacks usually lead to unrecognizable representations that are not useful to attackers in reality. 

To address this problem,  Basu {\em et al.}~\cite{basu2019membership} proposed another white-box attack method, where an attacker is able to access the target model and know the domain information of the target model ({\em e.g.}, the target model is trained on a human face or optical character recognition). 
With the domain information, an attacker can establish a GAN model to search correct representations in a quite low-dimension space by the generator $G$, formulated in Eq.~\eqref{model_inversion_eq}.
\begin{equation}
	\hat{z}=\arg \min_z {L}(f(G(z)),y)+\lambda R(z)
	\label{model_inversion_eq},
\end{equation}
where ${L}$ is the loss function of the target model, $\lambda$ is a parameter, and $R(\cdot)$ is a regularization term.
When an attacker learns the domain information, he can grab sufficient data in that domain as the training data from public data sources, such as Internet.
The grabbed data is used as the real dataset to train a traditional GAN model. 
After being trained, the generator $G$ is used in Eq.~\eqref{model_inversion_eq} to obtain an optimal low-dimension input $z$. 
Essentially, GAN acts as a transmitter to transfer a high-dimension problem into a low-dimension problem. 
Then the model inversion result can be generated with $\hat{x}=G(\hat{z})$ quickly. 
This method's attack efficiency is impressive as shown by the authors, and does solve the problem of unrecognized representations. 

Later, an improved version of model inversion attacks was developed in~\cite{Yuheng2019The}: Generative Model Inversion (GMI) that is similar to ~\cite{basu2019membership} but more powerful. 
The major merits of GMI lie in two aspects: 
(i) it can perform inversion attacks successfully even if the distribution of the attacker's prior information is different from that of the training dataset;
and (ii) an improved objective function makes attackers stronger. 
The implementation of GMI has two phases: {\em public information distillation} and {\em secret revelation}.
In the first phase, an attacker trains a WGAN model on public information so that the trained generator can be used to recover realistic data.
In the second phase, an attacker optimizes the latent vector $z$ ({\em i.e.}, the input of the generator) via $\hat{z}=\arg \min_z L_{prior}(z)+\lambda L_{id}(z)$, where $L_{prior}(z)$ is used to penalize unrealistic data, and $L_{id}(z)$ encourages the generated images to have maximum likelihood under the target model.
Additionally, it has been proved that the more accurate a model is, the easier it is to be attacked, which indicates a trade-off between accuracy and security vulnerability for a learning model.

\subsubsection{Protection of Preimage Privacy}
To preserve preimage privacy in the MLaaS scenarios, GAN-based mechanisms have been utilized to preprocess private data before the training stage. 
As shown in Fig.~\ref{fig:CPGAN}, Compressive Privacy Generative Adversarial Network (CPGAN)~\cite{tseng2020compressive} contains three modules: 
(i) the generator $G$ that is a privatization mechanism for generating privacy-preserving data;
(ii) the service module $S$ providing prediction service with the predicted label $Y$ as utility; 
and (iii) the attacker module $A$ that is a mimic attacker aiming at getting the reconstructed data $X'$ using $Z$.
More specifically, producing $Z$ in $G$ requires that the prediction service $S(Z)$ should perform well and the reconstruction error of $A(Z)$ should be large even if the attacker $A$ is the strongest, based on which the objective of CPGAN can be formulated by Eq.~\eqref{eq:cpgan}.
\begin{equation}
	\label{eq:cpgan}
	\max_G[\min_A L_A (X,A(G(Z)))-\lambda \min_S L_S (S(G(X)),Y)].
\end{equation}
CPGAN can defend preimage privacy attacks in MLaaS because the input data of $S$ does not contain any sensitive information.
However, the generator $G$ directly accesses sensitive data, which introduces potential privacy threats to $G$.
Furthermore, the un-stable training property of GAN makes the optimization process hard to converge.
\begin{figure}
	\centering
	\includegraphics[width=0.6\linewidth]{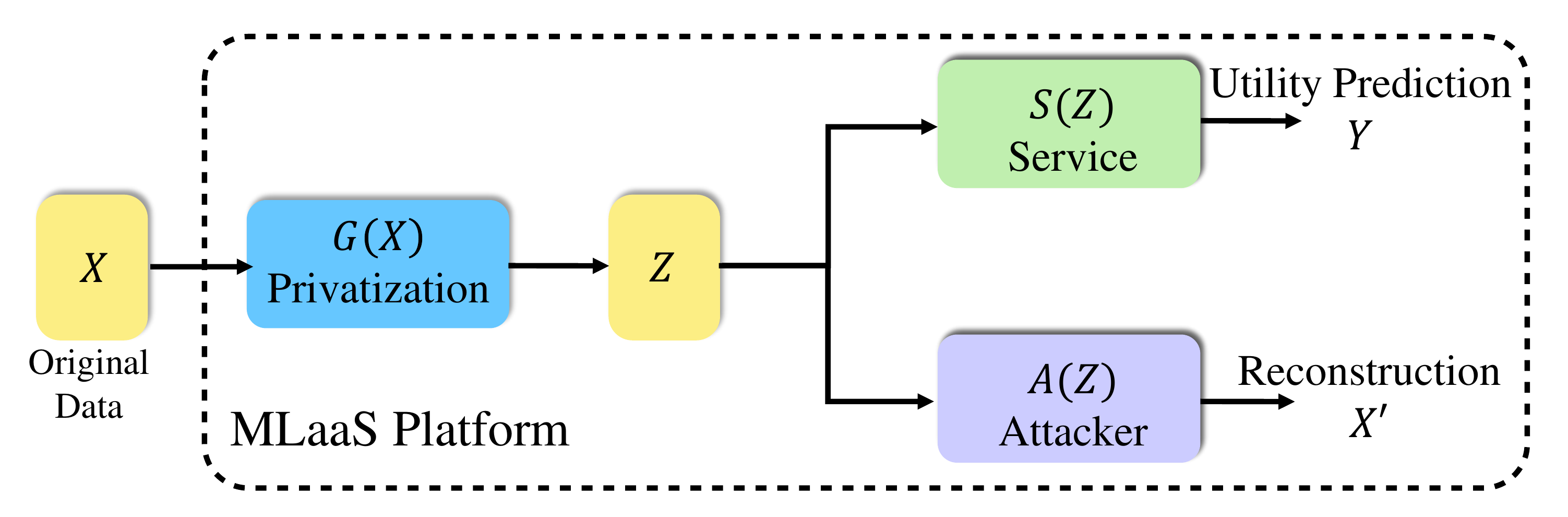}
	\caption{The Framework of CPGAN.}
	\label{fig:CPGAN}
\end{figure}

\subsection{Privacy in Distributed Learning Systems}
\label{sec:distributed_privacy}

As aforementioned, {\em membership privacy} and {\em preimage privacy} can be maliciously inferred in an end-to-end centralized machine learning system~\cite{ateniese2013hacking, fredrikson2014privacy} because all sensitive information is held by a central server that is likely to suffer single point failure. 
To address this issue, decentralized learning systems become promising solutions, in which the geographically distributed data is trained by different participants locally without data sharing.
Two most popular distributed learning schemes are distributed selective SGD (DSSGD)~\cite{shokri2015privacy} and federated learning (FL)~\cite{mcmahan2017communication}.
In DSSGD, local models only share and exchange a small fraction of parameters through a remote server. 
In FL, the server aggregates parameters using the submitted locally-trained parameters.
Both methods can train a global model built on a server with an accuracy comparable to that in the centralized learning system~\cite{shokri2015privacy}. 
Although distributed learning can protect data privacy to some extent as no one could have a global view of all training data, it is still far from perfect.

\subsubsection{Privacy Attacks in Distributed Learning Systems}
Since generator of GAN can mimic data distribution, a well designed GAN-based model can threaten data privacy in distributed learning scenarios. 
Besides membership privacy and preimage privacy, more challenging privacy issues should be handled for distributed learning. For example, attackers can invade a server or local users to steal parameters; more private information needs to be protected, {\em e.g.}, which data belongs to which local users or which users participate in the distributed training process; and how to identify/defend malicious servers or local users who pretend to be trusted. 

The first attack method targeting DSSGD was proposed by Hitaj {\em et al.}~\cite{hitaj2017deep} using GAN as an attacker, in which an attacker pretends to be an honest local model within the distributed training system with a goal of recovering sensitive information of a specific label that he does not have. 
Without compromising the central server or any local models, the attacker only uses parameters shared by other models and some common information ({\em i.e.,} all class labels in a training dataset) to build a local dynamic GAN.
Unlike model inversion attacks, the attacker can update its GAN in real-time to adjust attack performance as long as the entire training is not stopped.
The attacker's key principle is to share a crafted gradient to the central server to push a victim model to upload more local data information.

In the distributed learning systems, the central server may be untrusted as well.
As studied in~\cite{wang2019beyond}, user-level privacy in the distributed learning systems can be revealed invisibly by the malicious server via training a multi-task GAN with auxiliary identification (mGAN-AI) without affecting system performance.
In mGAN-AI, $G$ is a conditional generator outputting fake data with random noise and data label as input, and $D$ is a multi-task classifier built from the shared model in the distributed learning systems.
When $D$ is under training, except for the last layer, the shared model with additional three parallel fully connected layers is copied to $D$ for the purposes of data generation, categorization, and identification.
To find an optimal solution, $G$ is trained by minimizing the loss of data generation, classification and victim identification, while $D$ is trained by maximizing the loss of data generation and victim identification.
After the training process, the attacker can use $G$ to generate sensitive information of any target victim model.
Besides, a more powerful active attack is provided, in which the malicious server allocates an isolated model to a victim model without sharing any model. 
In this case, $D$ in the attacker's GAN model is exactly the same as the model uploaded by the victim.
As a result, the attacker can reconstruct $G$ to produce more accurate data without influence on other local models.
But this active model actually affects the protocol of distributed learning and decreases learning performance compared with original attacks.
Moreover, there is a very strong assumption: the server has a global dataset. 
If this is true, there is no need to use GAN for attack implementation.

\subsubsection{Privacy Protection in Distributed Learning Systems}
Privacy protection in distributed learning systems also deserves our attention.
Yan {\em et~al.}~\cite{yan2019amethod} designed a protection mechanism against two different attack behaviors, including stealing user information and attacking server parameters.
In their protection mechanism, except for the attacker, every local model is embedded with an additional layer called ``buried point layer'' and all its weights are set to be $0$.
When an attacker starts to attack, for the sake of an unknown ``buried point layer'', the parameters uploaded to the server should be different from harmless local models.
At the server, a detection module is used to detect abnormal changes.
If the attacker uploads parameters to the server, the detection module immediately discovers the intrusion. 
When an intrusion is detected for the first time, the link between the attacker and the server is awaited for a check; and when an intrusion is detected for the second time, the connection is blacklisted.

A few current works focus on the extension of GAN to a federated scenario. 
Due to the constraint that no raw data can leave its local dataset, federated learning is somehow restricted to train classifiers only and thus cannot be used in other important applications, such as data generation and reinforcement learning, especially on small datasets.
The integration of GAN and federated learning can realize distributed data generation, improving traditional federated learning's applicability.
Generally speaking, federated GAN's objective is to obtain a global generator at the server to produce realistic data following the data distribution of local clients without privacy leakage.
In~\cite{hardy2019md}, the generated data and corresponding errors are exchanged for data generation among a generator at the server and distributed discriminators at local clients.
Similar updating rules are adopted in~\cite{yonetani2019decentralized} for data generation in a non-i.i.d. setting by assigning different weights to local discriminators at the aggregation stage.
Rasouli {\em et al.}~\cite{rasouli2020fedgan} built the federated GAN in another way that trains both the generator and the discriminator locally on each private dataset and employs the server as a parameter aggregator and distributor.

\subsection{Differential Privacy in GAN}

According to the analysis in Sections~\ref{sec:member:privacy}-\ref{sec:distributed_privacy} and the conclusion of~\cite{song2017machine}, we can find that the root cause of privacy leakage of models is that machine learning models remember too much. That is, during the training process of a model, the model parameters are optimized to fit the underlying training dataset, which implies that the information of training data ({\em e.g.,} distribution, features, membership, {\em etc.}) is embedded into the model parameters.
Therefore, the adversary can unveil private information by exploiting the model parameters.

So far, two types of solutions have been proposed to overcome this vulnerability in learning models. 
The first method is adding a regularization term in a loss function to avoid overfitting during the training process. 
The regularization item can improve robustness and generalize a model to work on the data that the model has never seen.
For example, this method can be used to defend membership inference attack~\cite{nasr2018machine} as described in Section~\ref{sec:protect:membership}. 
The second method is to add acceptable noise into model parameters to hinder privacy inference attacks. 
Such a kind of obfuscation method ({\em e.g.}, $k$-anonymity, $l$-diversity, $t$-closeness, and differential privacy) has attracted lots of research interests for privacy protection, especially the combination of differential privacy~\cite{chaudhuri2019capacity} and neural networks~\cite{abadi2016deep}.
Notably, the recent research~\cite{yeom2018privacy, wu2019generalization} have illustrated the relation between differential privacy and the over-fitting problem: introducing differential privacy noise into model parameters could reduce over-fitting, thereby mitigating the privacy leakage.

Acs {\em et al.}~\cite{acs2018differentially} presented a first-of-its-kind attempt to build private generative models based on GAN.
In their method, GAN is trained to generate unlimited data for data release with a differential privacy guarantee, which improves the generative model's performance significantly and eliminates the constraints of limited data sources.
The proposed method divides the whole training dataset into $k$ disjoint sub-datasets using the differentially private $k$-means algorithm and trains the local generative models on each sub-dataset separately.
In~\cite{jordon2018pate}, a PATE mechanism-based model named PATE-GAN also adopted a dataset dividing strategy.
In PATE-GAN, there is a generator $G$, a teacher $T$ with $k$ teachers trained on $k$ disjoint datasets, and a student $S$. 
First, $T$ is trained with public data to differentiate real/fake as the discriminator.
Then $T$ is used as a noisy label generator to produce a differentially private dataset for training student $S$ that is the real discriminator.
Especially, $G$ and $S$ work as a couple of networks to generating realistic data with a privacy guarantee. 
The noticeable thing is that during the adversarial training of GAN, $T$ is also updated with $G$ and $S$ at each iteration, which is better for the discrimination capability of $T$.  
The training performance of the above approaches is dependent on the value of $k$.
If an appropriate $k$ is chosen, the training process would be benefited; otherwise, the training process would suffer because an inappropriate $k$ is possible to induce uneven clustering and a large privacy budget.
To get rid of the restriction of $k$, Xie {\em et al.}~\cite{xie2018differentially} and Liu {\em et al.}~\cite{liu2019ppgan} proposed another more straightforward differentially private GAN model (DPGAN) by carefully adding designed noise into gradients during the training procedure.
Based on the training process of GAN, DPGAN makes several improvements for privacy protection.
First, the loss function of WGAN is adopted to generate better results and resist model collapse, which is crucial for considering a trade-off between privacy and utility.
Then, at the training stage, Gaussian noise is added into gradient calculation.
After that, the updated weights are clipped by an upper bound, which helps achieve a smaller privacy loss.
Since the generator has no way to access the original data, there is a waste of privacy budget for adding noise on the generator.
Thus, the privacy loss of DPGAN can be further reduced.

\begin{table}
	\caption{Comparison of GAN-based Mechanisms for Model Privacy Protection.}
	\label{tab:model_privacy}
	\resizebox{\linewidth}{!}{
		\begin{tabular}{l c c c c c c}
			\toprule
			Literature & Purpose & White/Black Box & Scenario & GAN Model & Requirement & Model Privacy \\
			\midrule
			\cite{hayes2019logan} & Attack & Both & MLaaS & GAN & Partial training data & Membership  \\
			\cite{Liu2018PerformingCA} & Attack & White & Centralized & GAN & Partial training data & Membership  \\
			\cite{nasr2018machine} & Protection & - & MLaaS & GAN+Regularization & large amount of data & Membership  \\
			\cite{wu2019generalization} & Protection & - & Centralized & WGAN & large amount of data & Membership  \\
			\cite{Avodji2019GAMINAA} & Attack & Black & MLaas/Centralized & BEGAN & Multiple query+random dataset & Preimage  \\
			\cite{basu2019membership} & Attack & White & Centralized & GAN & Domain information & Preimage  \\
			\cite{Yuheng2019The} & Attack & White & Centralized & WGAN & public information & Preimage  \\
			\cite{tseng2020compressive} & Protection & - & MLaaS & GAN & large amount of data & Preimage  \\
			\cite{hitaj2017deep} & Attack & White & Decentralized & GAN & target label & Data feature  \\
			\cite{wang2019beyond} & Attack & White & Decentralized & cGAN & Malicious server & User information  \\
			\cite{hardy2019md,yonetani2019decentralized,rasouli2020fedgan} & Protection & - & Decentralized & GAN & Local training data & Local data privacy \\
			\cite{acs2018differentially,jordon2018pate} & Protection & - & Centralized & GAN & $k$, public data & Membership  \\
			\cite{xie2018differentially,liu2019ppgan} & Protection & - & Centralized & WGAN & DP & Data feature  \\
			\cite{zhang2018differentially,xu2019ganobfuscator} & Protection & - & Centralized & WGAN & Public dataset, DP & Data feature  \\
			\cite{triastcyn2018generating} & Protection & - & Centralized & cGAN & target labels, DP & Membership  \\
			\cite{triastcyn2020federated,augenstein2019generative} & Protection & - & Decentralized & GAN & DP & Data feature  \\
			\cite{lu2019empirical} & Protection & - & Centralized & GAN & New defined metrics & Data feature  \\
			\bottomrule
		\end{tabular}
	}
\end{table}

To this end, Zhang {\em et al.}~\cite{zhang2018differentially} and Xu {\em et al.}~\cite{xu2019ganobfuscator} proposed two methods adding differential noise on the discriminator only. 
The basic idea of these two methods are too similar, so we review them here together.
At the beginning of training, WGAN is used for pursing stability of the training process.
During the training procedure, the gradients of the discriminator are bounded and perturbed using the methods of DPGAN~\cite{xie2018differentially}.
However, the generated data has low quality, and the proposed models converge slower than the traditional GAN, resulting in excessive privacy loss.
The following three solutions were proposed to improve data quality, convergence rate, training stability, and scalability. 
(i) parameter grouping is a common scheme used in differential privacy~\cite{han2018research}. 
A balance between convergence rate and privacy cost could be achieved by carefully grouping the training parameters and clipping over different groups.
(ii) adaptive clipping can enhance the data quality of using random clipping. 
Assume that a public dataset can be used to dynamically adjust the clipping bounds to achieve faster convergence and stronger privacy protection.
(iii) warm starting can save privacy budget for critical iterations, in which a public dataset is used to initialize the models with a good starting point.
This work's shortage is that the assumption of an available public dataset is too strong in such a private data release scenario.
Until here, all the generative models with differential privacy focus on unlabeled data, and the work on labeled data has not been addressed.

In~\cite{triastcyn2018generating}, the authors developed a cGAN-based framework for the generation of labeled data.
The proposed framework can achieve differential privacy for both the generator and the discriminator by only injecting noise into the discriminator.
This framework's advantage is that with the class label as auxiliary information in cGAN, the quality of generated data is extremely high with a low privacy loss.
	However, since the generative model observes the label information, more information may be leaked to attackers when the model is published. 
As a result, this framework might only be able to prevent membership inference attacks but fail to protect preimage privacy.

As discussed in Section~\ref{sec:distributed_privacy}, for the distributed GAN, the privacy of generators in federated setting also needs to be protected because the physical separation of data is not enough.
More analysis of differential privacy in the private federated GAN models was presented in~\cite{triastcyn2020federated,augenstein2019generative}, where the authors proposed to add differential privacy during the training process of local GAN models. 
The noise mechanisms also utilize clipping and noisy gradients, which is similar to the methods in~\cite{xie2018differentially, zhang2018differentially, xu2019ganobfuscator}.
To achieve a better trade-off between utility and privacy, a relaxed expected privacy loss is adopted by~\cite{triastcyn2020federated}, while the work in~\cite{augenstein2019generative} utilizes the original {\em Moments Accountant} strategy. 
In a nutshell, differential privacy is a classic standard for privacy protection and an efficient approach for preserving the membership and preimage privacy of GAN.
But the protection efficacy of differential privacy is greatly determined by the noise scale, which may introduce utility loss and needs more research endeavors.

Not stopping here, a study performed by Lu {\em et al.}~\cite{lu2019empirical} pointed out that even there had been many researches on private data release by generative models, the adopted quality metrics are not quite suitable.
For example, 
adopting differential privacy may impair data utility more or less. 
Thus the users ({\em e.g.}, companies) often do not adopt the strict privacy guarantee in academic areas. 
Instead, they only require the privacy level of some mechanisms to be slightly better than the government regulation even in the boundary of law. 
For those users, a more practical measurement could be considered.
In~\cite{lu2019empirical}, the experiments were set up to evaluate new defined metrics in addition to the privacy budget of differential privacy {\em e.g.,} hitting rate, record linkage and Euclidean distance.
The extensive experiments demonstrate that for formal privacy definition ({\em i.e.}, differential privacy) even though it achieves strict privacy protection but losses more data utility.
For industrial and commercial applications, informal privacy guarantees, such as GAN-based methods, can meet privacy requirements and have a better data utility.

At the end, we compare the above reviewed methods in Table~\ref{tab:model_privacy} from different perspectives.

\section{Security with GAN}
\label{sec:security}
Besides privacy issues, a variety of security issues also exist in GAN.
In this section, the research findings on security for GAN are introduced from the aspects of model robustness, malware, fraud detection, vehicle security, industry protocol, {\em etc.} and a comparison is concluded in Table~\ref{tab:security_GAN}.

\subsection{Model Robustness}
For machine learning models, one of the most severe secure threats is adversarial sample attack. 
Let $X$ be the feature space of data and $Y$ be the class space.
Suppose for the original data $x\in X$, its ground-truth label is $y\in Y$.
For a given classifier $f: {X} \to {Y}$, an adversarial example attack intends to manipulate a sample $x^{\prime}$ through unperceptive modification to mislead classification, which can be mathematically formulated as the following optimization problem:
\begin{subequations}
	\begin{align}
		\label{eq:adv_sample_1a}       \min&~~L(x', x),\\
		\label{eq:adv_sample_1b}   \text{s.t.} &~~f(x')\neq f(x)=y. 
	\end{align}
\end{subequations}
Eq.~\eqref{eq:adv_sample_1a} implies the objective of minimizing the distance, $L(x^{\prime}, x)$, between the adversarial data and the original data, where $L(\cdot, \cdot)$ is a predefined distance metric ({\em e.g.}, $L_2$ norm).
Eq.~\eqref{eq:adv_sample_1b} indicates the incorrect classification result on the adversarial data.
There are two kinds of adversarial attacks: non-targeted and targeted.
Non-targeted attacks only require $f(x^{\prime}) \neq f(x)$, and targeted attacks expect $ f(x^{\prime}) = y_t \neq f(x)$ with $y_t \in Y$ being a target label pre-determined by an attacker.

\subsubsection{Adversarial Sample Attacks}
There have been many works on targeted attacks~\cite{moosavi2016deepfool, papernot2017practical} and non-targeted attacks~\cite{goodfellow2014explaining, szegedy2013intriguing} based on traditional optimization methods.
In~\cite{baluja2017adversarial}, Baluja and Fischer, for the first time, generated targeted adversarial samples using GAN to attack machine learning models.
Unlike previous works that produce adversarial samples by optimizing a noise $\delta$ added into the original data $x$, the idea of ~\cite{baluja2017adversarial} is to train a neural network to obtain an adversarial sample $x^{\prime}$ directly from the original data $x$.
Such an adversarial generator aims to simultaneously minimize the distance loss in the feature space and the classification loss in the prediction space.
Particularly, in the implementation of targeted attacks, an attacker hires a reranking function to resort to the predicted labels such that the target label has a maximum probability and the other labels maintain their original order. 
Compared with the traditional optimization-based methods, this generative attack mechanism is extremely fast and efficient once the neural network has been trained.
However, this attack mechanism is model-dependent, which means it does not perform well in the black-box scenario and lacks transferability.

To deal with the above weakness, Zhao {\em et al.}~\cite{zhao2017generating} suggested using a GAN-based model plus a data inverter to enhance attack capability.
They adopted WGAN to generate vivid data from random noise $z\sim \mathcal{N}(0,1)$ and used the dense representation $z$ to produce realistic adversarial samples, in which a data inverter $I$ was designed to map normal data to the corresponding dense representations $z\sim \mathcal{N}(0,1)$. 
With the help of this inverter $I$, any normal data can be transferred into its corresponding representation $I(x)$ used as input to produce an adversarial dense representation $\tilde{z}$ for sample generation in WGAN.
The work-flow of this attack is briefly summarized in Fig.~\ref{gan_inverter}.

\begin{figure}[h]
	\centering
	\includegraphics[width=0.6\linewidth]{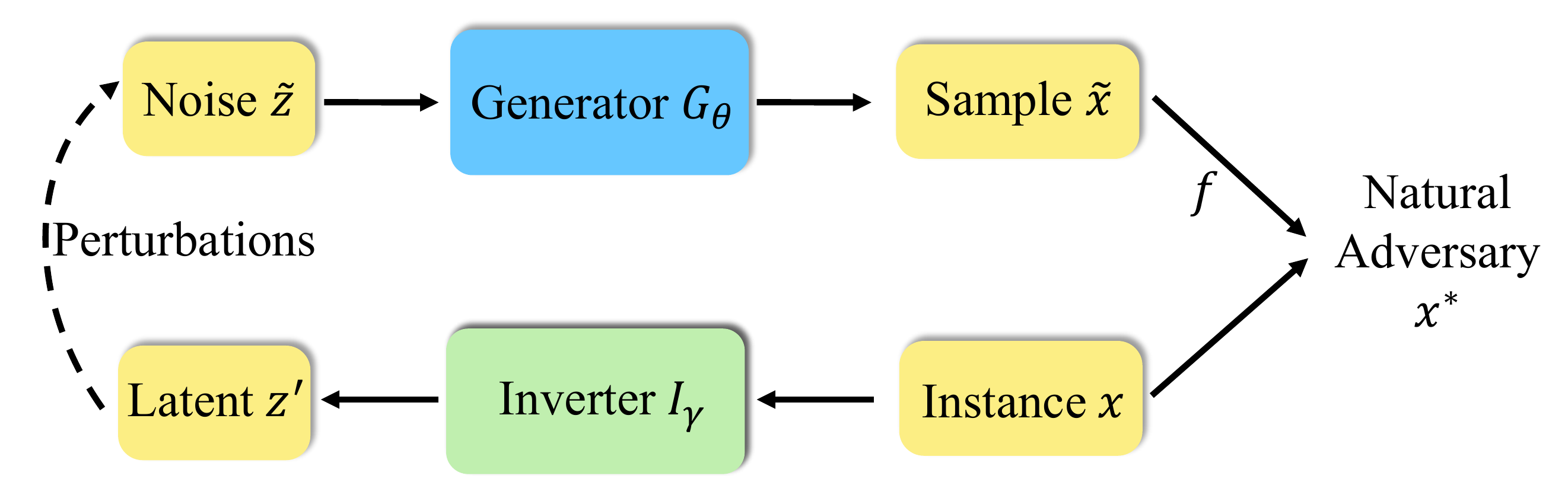}
	\caption{The Structure of GAN Inverter.}
	\label{gan_inverter}
\end{figure}

In~\cite{xiao2018generating}, a target classifier model was seamlessly integrated into GAN to train a stronger attacker in the black-box scenario.
The proposed model aims at minimizing the target prediction loss, normal form of GAN loss, and the noise scale simultaneously.
When performing a black-box attack, the whole network is updated to obtain a strong classifier $f$ via two steps: (i) fix the target classifier and update the generator and the discriminator in GAN according to the objective; and (ii) fix GAN and update the classifier satisfying $\arg \min_f \mathbb{E}_x H(f(x),b(x))+ H(f(x+G(x)),b(x+G(x)))$, where $b$ is the target black-box classifier, $f$ is a shadow model built from observation of $b$, and $H(\cdot, \cdot)$ is the cross entropy function.
After the training process, the generator in GAN can be used to produce adversarial sample $x+G(x)$.
Since the attack model is dynamically trained based on the target classifier's changes, the attack capability is enhanced, and the attack success rate is increased.

Unlike the previous attack methods requiring small adversarial noise for accurate human perception and classification, an unrestricted adversarial attack method was proposed in~\cite{song2018constructing} to mislead the victim classifier to the target label $y_t$ using a modified ACGAN model of~\cite{odena2017conditional}.
In the beginning, a generator $G(z,y)$ is trained to generate synthetic labeled data from random noise $z$ and data label $y$.
To generate the adversarial samples with target label $y_t$, an attacker tries to optimize $z$ to $\tilde{z}$ until $f(G(\tilde{z},y))=y_t$, where $f$ is a given victim classifier.
This attack is achieved by two optimization procedures in ACGAN: 
(i) for the victim classifier $f$, minimize the loss between $f(G(\tilde{z},y))$ and $y_t$; 
(ii) for the additional classifier $C$, minimize the loss between $C(G(\tilde{z},y))$ and $y$.
The advantage of this attack is that it can generate numerous adversarial samples that could be even different from the original data without being detected by a human.

Similar to~\cite{odena2017conditional}, Wang {\em et al.}~\cite{wang2019gan} designed an AT-GAN model using a pre-trained generator $G$ and ACGAN to transfer $G$ to $G_{attack}$ that can be used to produce adversarial samples directly.
In the beginning, an attacker trains $G(z,y)$ to mimic real data distribution with ACGAN.
Based on the training results of $G(z, y)$, the attacker slightly modifies $G(z, y)$ to $G_{attack}(z,y)$ by minimizing $\|G_{attack}(z,y)-G(z,y)\|$ such that $G_{attack}(z,y)$ can produce data with the target label $y_t$, where the prediction loss, $L(f(G(z,y)),y_t)$, and the distance loss, $\|G_{attack}(z,y)-G(z,y)\|$, are taken into account for simultaneous minimization.
For the desired adversarial samples, the prediction loss assures the modified samples can be classified to the target class, and the distance loss controls perturbation magnitude between real and fake data.

On the other hand, as a generative model, GAN itself also faces the danger of being attacked.
In~\cite{kos2018adversarial}, a framework was given to show how to design adversarial samples to attack generative models, such as GAN and VAE.
Apart from the original adversarial samples in classification models, an adversarial sample $x'$ in the generative model is achieved by minimizing the distance $L(x', x)$ such that $f(G(x'))=y_t \ne f(G(x))$.
To realize the attack purpose, the entire loss function, $\lambda L(x,x')+H(f(G(x')),y_t)$, should be minimized, in which $L(\cdot, \cdot)$ controls the distance between $x$ and $x'$, and $H(\cdot, \cdot)$ compels $G(x')$ to be classified to $y_t$ class.
This is the first work to design an attack model for the generative models.
However, the proposed attack model can only handle white-box attack implementation, making it infeasible to perform attacks without prior knowledge.

\subsubsection{Adversarial Sample Defense}
The existing strategies against adversarial sample attacks can be briefly classified into three categories: denoising, adversarial training, and detection.
The GAN-based defense methods will be summarized from the above categories.

The first denoising method based on GAN was designed by~\cite{shen2017ape} with the fundamental idea that the generator can take adversarial samples to output normal data.
Accordingly, a conditional GAN structure was established, where the generator $G(x')$ takes adversarial sample $x'$ as training data and learns the normal data distribution with $G(x')=x$.
That is, the distribution of the generated data should be the same as that of the real data to make the output become clean data.
In this GAN framework, the loss function of the discriminator is the same as that in the original GAN, while the generator's goal consists of a distance loss that constraints the distance between $x$ and $G(x')$ to be small and an adversarial loss that requires $D(G(x'))$ to have a higher score.
A similar idea was exploited by Samangouei {\em et al.} to develop a ``Defense-GAN'' framework that denoises adversarial samples to obtain normal samples through the generator of GAN~\cite{samangouei2018defense}.
The slight difference is that in~\cite{samangouei2018defense}, the denoising generator is trained on clean data from noise $z$ to minimize the distance between input data and generated data.
After the optimized $z^*$ is obtained, a reconstructed data $G(z^*)$ is fed into $f$ expecting that the output $f(G(z^*))$ is the same as normal data.
The denoising method can be used in conjunction with any classifier and does not need to modify the classifier structure. Thus it will not decrease the performance of the trained model.
Also, it is independent of any attack method, which means it can be used as a defense against any attack.

To tackle the issue of insufficient adversarial data in adversarial training, a generative adversarial training method was proposed by using GAN~\cite{lee2017generative}.
In this method, the generator $G(\nabla)$ takes the gradient $\nabla$ of normal data $x$ as input and outputs adversarial noise to perturb $x$. 
The loss function of $G$ is $L_G(\nabla,y)=H(f(x+G(\nabla)),y)+\lambda\|G(\nabla)\|^2$, where $H(f(x+G(\nabla)),y)$ encourages the generated data $x+G(\nabla)$ to be classified correctly by the classifier $f$, and $\lambda\|G(\nabla)\|^2$ requests the generated noise $G(\nabla)$ to be small and imperceptible.
The desired classifier $f$ was configured as a GAN's discriminator to reduce the classification loss for both the normal data and the generated data.
The loss function of the classifier $f$ is $L_f=\alpha H(f(x),y)+(1-\alpha)H(f(x+G(\nabla)),y)$, where $H(\cdot, \cdot)$ is defined as cross entropy.
This paper is the first work to apply GAN to adversarial training and provides a robust classifier so that enough adversarial data can be used to enhance the desired model's regularization power effectively.
It is worth pointing out that the proposed method is model-dependent and can only work on some specific models due to the classifier $f$'s involvement in the adversarial training.

An improvement has been made by Liu {\em et al.}~\cite{liu2019gandef} who proposed a model-independent method named ``GanDef'' that can be a defense for many different classifiers.
According to their analysis, the misclassification of deep models is caused before the soft-max layer.
The input of soft-max layer would be different from adversarial samples and normal samples through forward propagation, which explains why the final output of deep models is different.
Moreover, the normal samples hold a property of ``invariant features''; that is, if a set of data is classified into correct classes, the input of soft-max should follow the same distribution. 
The failure of classification on the adversarial samples indicates the adversarial samples and the normal samples do not have the same invariant features.
Thus, for correct classification, a classifier and a discriminator are deployed in a GAN framework to modify adversarial samples such that their invariant features are the same as normal data.
Following the training process of the original GAN, the classifier and the discriminator are iteratively updated.
Finally, the classifier is supposed to have the ability to make the adversarial sample's invariant features similar to (ideally, even the same as) the normal samples.
Thus, any data can be processed before soft-max layer with this mechanism and get the correct prediction.

\subsection{Malware Detection}
Hu \textit{et al.}~\cite{hu2017generating} developed MalGAN to generate adversarial malware examples, where GAN was employed as a binary malware feature generator to attack a malware detector.
The malware only extracts the program output of detection with the assumption that the attacker only knows the detector's features without the machine learning algorithm the detector uses and the parameters of the trained model. 
As the trained model details are unkown, a substituted detector is used to fit the black-box detector and provide gradient information.
The adopted substituted detector's training data has two parts: the set of adversarial malware examples from the generator of GAN and the set of examples from an additional benign dataset. 

In~\cite{shahpasand2019adversarial}, GAN was applied for detecting Android malware. 
The detection focuses on black-box attacks, where attackers cannot access the inner details of the network (including network architecture and parameters) but can get the classifier's output and alter the malware codes based on detection results. 
Generally, given a malware $x$ with the true label $f(x)=1$, the attacker aims to avoiding malware detection via generating an adversarial version $x' = x + \delta$ such that $f(x')=0$, where $f( \cdot )$ is a malware detector.
The loss function includes the similarity between the generated and the benign samples (denoted by $L_{GAN}$) and misclassification rate of the adversarial malware samples (denoted by $L_{adv.Mal}$), which can be mathematically expressed as: 
$L=\alpha L_{GAN}+(1-\alpha)L_{adv.Mal}$. 
Where the $L_{GAN}$ represents discriminative loss of GAN on real and fake data and $L_{adv.Mal}= l_{f}(x+G(z), 0)$ is supposed that an adversarial sample can bypass the targeted classifier $f$ by manipulating itself as a benign class.

Furthermore, GAN can be utilized for analyzing Linux and Windows malware. 
Kargaard \textit{et al.}~\cite{kargaard2018defending} brought GAN to the analysis of malware detection, where the malware binaries are converted to images for training in GAN.
In particular, the malware is collected via a honeypots system and contains WannaCry ransomware, Linux SMB trojan, and MySQL Trojan, {\em etc}. 
All the malware binary files are converted to greyscale images with a size of $32 \times 32$ for processing.
Rahim \textit{et al.}~\cite{9288942} proposed using Federated GAN to defend attack and enable the devices to communicate with each other efficiently and securely.
As illustrated in this work, the proposed method is much more effective and reliable than the previous methods.

Among various malware, zero-day malware is outstandingly difficult to be detected because it cannot be removed by antivirus systems that mainly use the characteristics of stored malware for detection. 
Kim \textit{et al.}~\cite{kim2017malware, kim2018zero} proposed a transferred deep-convolutional generative adversarial network (tDCGAN) that applies a deep autoencoder to learn malware characteristics and transfers the characteristics to train the generator of GAN. 
The architecture of tDCGAN has three parts: 
(i) data compression and reconstruction, in which preprocessed data is input to compress and reconstruct the malware data;
(ii) fake malware generation, in which deep autoencoder is used to reconstruct malware data, and the decoder is transferred to the generator of GAN;
and (iii) malware detection, in which the generator is given the probability distribution to produce fake malware. 
Then the discriminator of GAN is transferred to the malware detector.

\subsection{Bioinformatic-based Recognition}

Bio-information ({\em e.g.}, fingerprints and iris) recognition systems have been widely deployed in many areas, such as banking, criminal investigation, and national security.
However, the bio-information gathering process is expensive and time-consuming. 
Also, due to the privacy protection legislation, publishing a bio-information database is not easy. 
Fortunately, GAN provides a novel way to construct bio-information systems for authentication.

Bontrager \textit{et al.}~\cite{bontrager2017deepmasterprint} used GAN to generate synthetic fingerprints for a fingerprint verification system identifying different people. In their work, two methods are designed based on GAN, with the first one applying evolutionary optimization in the space of latent variables and the second one using gradient-based search. 
In~\cite{kim2019fingerprint}, the fingerprints, namely the master minutia set, were generated from a two-stage GAN. 
The two-stage GAN is composed of two GANs: the first GAN is for generally describing fingerprints, and the second GAN uses the outputs from the first GAN to create fingerprint images. 
Then the minutiae extracted from the second GAN's outputs and a feature extraction algorithm is used to do extraction. 

\begin{figure}[tb]
	\centering
	\subfigure[Makeup Transfer Sub-network]{
		\begin{minipage}[t]{0.4\linewidth}
			\centering
			\label{Subnet1}
			\includegraphics[width=\linewidth]{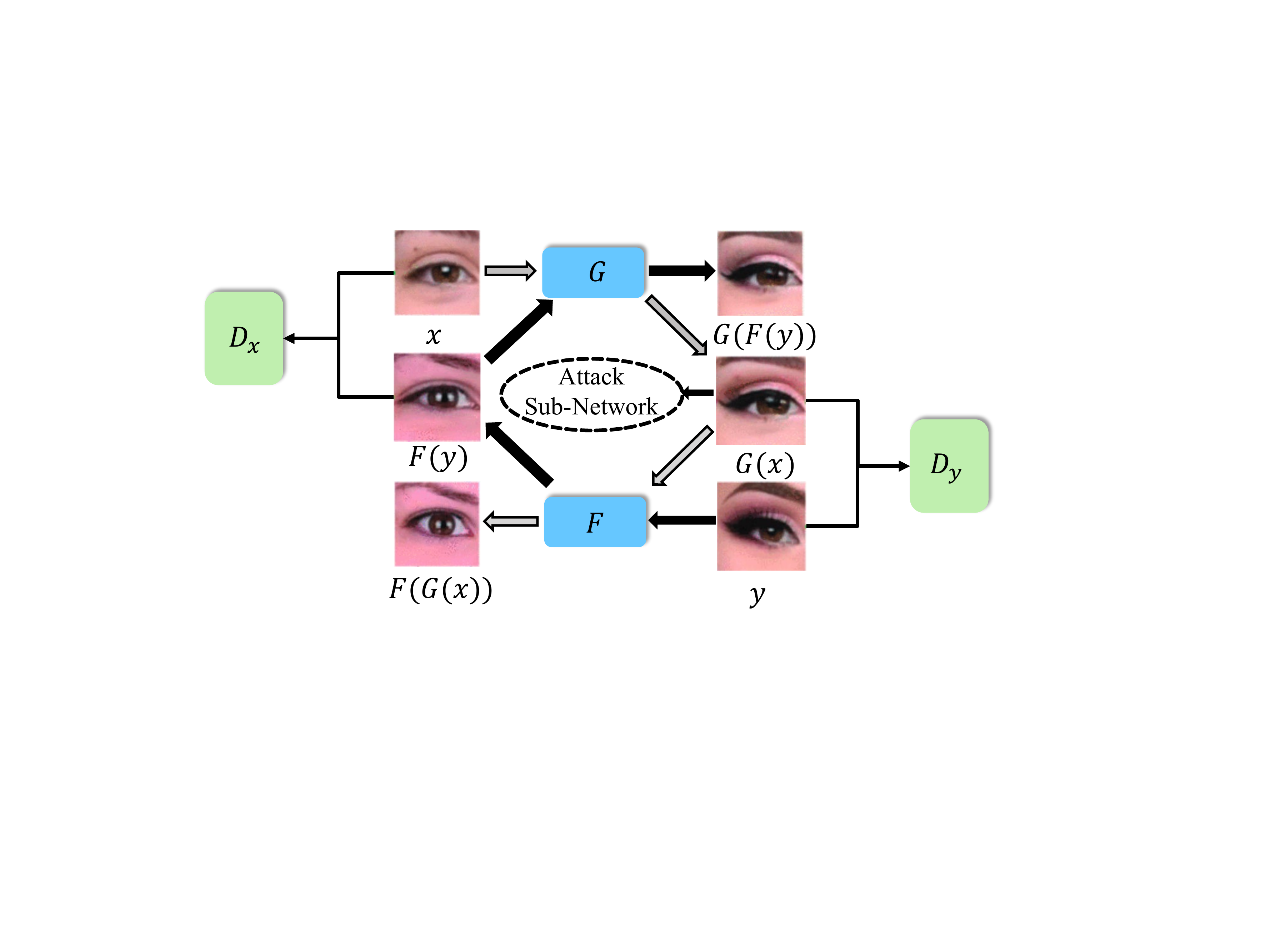}
		\end{minipage}%
	}%
	\subfigure[Adversarial Attack Sub-network]{
		\begin{minipage}[t]{0.38\linewidth}
			\centering
			\label{Subnet2}
			\includegraphics[width=\linewidth]{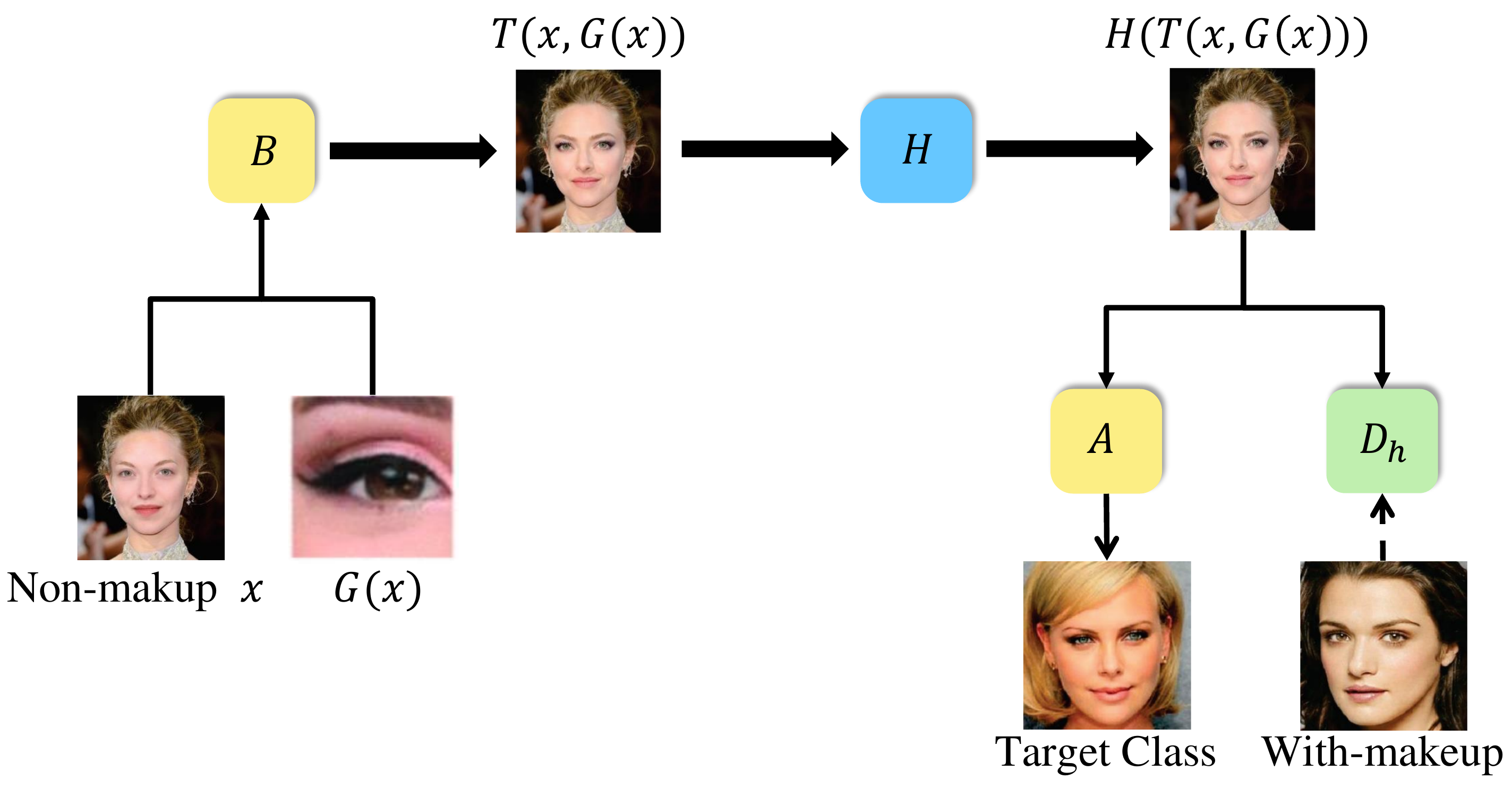}
		\end{minipage}%
	}%
	\caption{The Framework of Adversarial Attack Networks. Image Source (http://iprobe.cse.msu.edu/datasets.php).}
\end{figure}
	For face recognition, there are also more and more attack methods that generate adversarial examples to violate the deep learning based face recognition models.
	Zhu \textit{et al.}~\cite{zhu2019generating} applied GAN to make up adversarial example attack on well-trained face recognition models. 
	The proposed method consists of two GAN-based sub-networks, including a ``make up transfer sub-network'' that transfers face images from non-makeup domain to makeup domain and an ``adversarial attack sub-network'' that generates adversarial examples. 
	The configuration of ``make up transfer sub-network'' follows CycleGAN~\cite{zhu2017unpaired} as shown in Fig.~\ref{Subnet1}, in which $x$ is the real non-makeup input, $y$ is the real makeup input, the generator $G$ outputs $G(x)$ to obfuscate the discriminator $D_y$, and the network $F: Y\rightarrow X$, generates results of non-makeup faces to obfuscate the discriminator $D_x$. 
	In the ``make up transfer sub-network'', one generator adds makeup effect to non-makeup images, and the other removes makeup effect while still maintaining the original identity. 
	On the other hand, $D_x$ distinguishes between the real non-makeup photos and the generated ones, and $D_y$ distinguishes between the real makeup photos and the generated ones. 
	The structure of ``adversarial attack sub-networks'' is presented in Fig.~\ref{Subnet2}, where a transformation $T(\cdot, \cdot)$ is input with the blend of original image $x$ and the generated output $G(x)$ of ``makeup transfer sub-network''.
	Then $T(x,G(x))$ serves as the input of the generator $H$, which is used to generate images $H(T(x,G(x)))$ that can deceive both the target network $A$ and the discriminator $D_h$.
	In the ``adversarial attack sub-networks'', the discriminator $D_h$ functions similarly as $D_y$, and $D_y$ is also used as a pre-trained model parameter to initialize $D_h$.

	Besides, Damer \textit{et al.} proposed MorGAN~\cite{8698563} to launch a realistic morphing attack by considering the representation loss.
	MorGAN is inspired by the idea of adversarially learned inference but uses a variational autoencoder instead of simple autoencoders.
	By doing so, MorGAN can avoid the possibilities of non-continuous latent space and further lead to more realistic output from the interpolation between encoded vectors.

\subsection{Fraud Detection}

The financial fraud detection problems, including credit card fraud, telecom fraud, and insurance fraud, {\em etc}., are well-known for the highly nonlinear and complex solutions. 
Artificial neural network (ANN), which simulates interacting neurons' properties, has been successfully utilized to solve such problems. 
However, in real applications for fraud detection, ANN faces two problems:
(i) the receiving bank cannot get access to detailed information about the sending accounts when the transaction happens between different banks;
and (ii) the receiving bank also cannot obtain call records of the recipients of transfers from the telecommunication provider.

To deal with these problems, in~\cite{zheng2018generative}, GAN was applied to telecom fraud detection under the bank receiving scenario, which is also useful for many other anomaly detection problems when the training dataset is limited.
In this work, the authors coupled GAN, autoencoder, and Gaussian Mixture Models (GMMs) for fraud detection. 
Particularly, the encoder together with a GMM was used as the discriminator, and decoder was used as the generator. 
Then the encoder and another GMM output the classification results, {\em i.e.}, whether a given sample is normal or fraud.

\subsection{Botnet Detection}

As one of the most formidable threats to cybersecurity, a botnet is often enrolled in launching large-scale attack sabotage~\cite{feily2009survey}.
The network-based detection mainly studies the abnormal characteristics of botnets based on network flows. 
Some other classic detection approaches are based on the extraction and selection of features using statistical analysis, machine learning, data mining, and other methods. 
However, these traditional detection schemes have two main shortcomings.
On the one hand, most of the existing network-based methods for botnet detection are limited to the packet inspection level and focus on partial characteristics of network flows, which cannot fully characterize botnets' abnormal behaviors.
On the other hand, botnets keep pace with the times and take advantage of advanced ideas and technologies to escape detection, raising insurmountable challenges to the traditional detection schemes.

The discriminator is essentially a binary classifier that classifies the samples into real or fake categories. 
Similarly, the real samples can be further categorized into normal traffic or abnormal traffic for a botnet detector.
Inspired by these observations, Yin \textit{et al.}~\cite{yin2018enhancing} proposed a GAN-based botnet detection framework, which is suitable for augmenting the original detection model.  
In~\cite{yin2018enhancing}, the discriminator was replaced with a botnet detector, and the corresponding binary output ({\em i.e.}, normal and anomaly) was transformed into a triple output ({\em i.e.}, normal, anomaly, and fake) for detection using the softmax function.

\subsection{Network Intrusion Detection}

Network environment is very complex and time-varying, so it is difficult to use traditional methods to extract accurate features of intrusion behaviors from the high-dimensional data samples and process the high-volume data efficiently. 
Even worse, the network intrusion samples are submerged into many normal data packets, leading to insufficient samples for model training. 

Yang \textit{et al.}~\cite{yang2019simple} proposed a DCGAN-based method to extract features directly from the raw data and then generated new training datasets by learning from the raw data, in which long short-term memory (LSTM)~\cite{gers1999learning} was applied to learn the features of network intrusion behaviors automatically.
The generator $G$ was configured with CNN, where the pooling layer was replaced with the fractional stride convolutions. 
Then, the fully connected layer was removed, ReLU was applied for all layers except the output layer, and tanh is used at the output layer. 
In addition, batchnorm was utilized to solve the poor initialization problem and propagate each layer's gradient. 
The discriminator was also constructed using CNN that contains a pooling layer without any fully connected layer, and LeakyReLU is used at all layers.
Batchnorm was used to propagate the gradient to each layer to avoid the generator converging all samples to the same point.

\subsection{Vehicle Security}
Seo \textit{et al.}~\cite{seo2018gids} proposed a GAN-based Intrusion Detection System (GIDS) for vehicular networks. A Controller Area Network (CAN) bus in the networks is an efficient standard bus enabling communication between all Electronic Control Units (ECUs). However, CAN itself is vulnerable due to the lack of security features. GIDS aims at detecting unknown attacks on CAN with two discriminative models.
The first discriminator receives normal and abnormal CAN images extracted from the actual vehicle. Because the first discriminator uses attack data in the training process, the type of attacks that can be detected may be limited to the attack used for training.
The generator and the second discriminator are trained simultaneously in an adversarial process, where the generator generates fake images by using random noise and the second discriminator determines whether its input images are real CAN images or fake images generated by the generator. 
In GIDS, the second discriminator ultimately beats the generator so that the second discriminator can detect even the fake images that are similar to real CAN images.
%

\subsection{Industry Protocols}
In industry, people often use fuzz to detect whether the industrial network protocols (INPs) are secure.
Traditionally, to generate the fuzzing data effectively, the guidance of protocol grammar is applied to the generating process, where the grammar is extracted from interpreting the protocol specifications and reversing engineering in network traces.

Hu \textit{et al.}'s work~\cite{hu2018ganfuzz} employed GAN to train the generation model on a set of real protocol messages for industrial network protocol fuzzing. 
Specifically, in the GAN framework, an RNN with LSTM cells is used as the generative model, and a CNN was set as the discriminative model. 
They used the trained generative model to produce fake messages, based on which an automatic fuzzing framework was built to test INPs. 
Their experiments showed that since the proposed framework does not rely on any specified protocols, the proposed framework outperforms many previous frameworks. 
Moreover, some errors and vulnerabilities were identified successfully in a test on several simulators of the Modbus-TCP protocol.

\begin{table}
	\caption{Comparison of GAN-based Mechanisms for Security.}
	\label{tab:security_GAN}
	\resizebox{\linewidth}{!}{
		\begin{tabular}{l c c c c c c}
			\toprule
			Literature & Purpose & White/Black Box & Application & GAN Model & Strategy & Target \\
			\midrule
			\cite{baluja2017adversarial} & Attack & White & Break robustness & GAN & Generating adversarial samples & Misclassification  \\
			\cite{zhao2017generating} & Attack & Black & Break robustness & WGAN & Explore latent space $z$ & Misclassification  \\
			\cite{xiao2018generating} & Attack & Black & Break robustness & GAN+shadow model & Minimize noise scale by classifier & Misclassification  \\
			\cite{song2018constructing} & Attack & White & Break robustness & ACGAN & Explore latent space $z$ & Misclassification  \\
			\cite{wang2019gan} & Attack & Black & Break robustness & ATGAN & Perturb normal generator & misclassification  \\
			\cite{kos2018adversarial} & Attack & White & Data generation & VAE-GAN & Modify latent representation & Incorrect generation  \\
			\cite{shen2017ape,samangouei2018defense} & Defense & Black & Data sanitizing & cGAN & Use generator to clean adversarial data & Denoising  \\
			\cite{lee2017generative} & Defense & White & Adversarial training & GAN & Generating adversarial samples & Enhance classifier  \\
			\cite{liu2019gandef} & Defense & White & Data generation & GAN & Modify invariant features in adversarial samples & Enhance classifier  \\
			\cite{hu2017generating} & Attack & Black & Malware detection & GAN & Generate adversarial malware examples & Invade detector  \\
			\cite{shahpasand2019adversarial} & Defense & Black & Malware detection & GAN & Generate noise for malware & Enhance detector  \\
			\cite{kargaard2018defending} & Defense & White & Malware detection & GAN & Convert malware into greyscale images & Enhance detector  \\
			\cite{kim2017malware, kim2018zero} & Attack & White & Malware detection & DCGAN & Generate malware & Invade detector  \\
			\cite{bontrager2017deepmasterprint,kim2019fingerprint} & Defense & White & Bioinfo recognition & GAN & Generate fingerprints & Bypass verification  \\
			\cite{zhu2019generating} & Attack & White & Bioinfo recognition & cycleGAN & Modify face without changing ID & Misclassification  \\
			\cite{zheng2018generative} & Defense & White & Fraud detection & GAN & Generate more fraud data for training & Enhance detector  \\
			\cite{yin2018enhancing} & Attack & White & Botnet detection & GAN & Modify discriminator to a detector & Enhance detector  \\
			\cite{yang2019simple,zhu2019generating} & Defense & White & Network intrusion detection & DCGAN & Generating intrusion behavior data & Data Augmentation  \\
			\cite{hu2018ganfuzz} & Defense & White & Industry protocols & GAN & Generate data for industrial protocol fuzzing & Data Augmentation  \\
			\bottomrule
		\end{tabular}
	}
\end{table}

\section{Future Works}
\label{sec:future_work}
The survey of the state-of-the-art GAN models displays the innovative contributions of GAN to solving the issues of privacy and security in various fields.
As a preliminary attempt, GAN's potentials have not been fully and deeply explored by the existing GAN-based approaches yet, leaving many unsolved challenging problems.
This section provides a comprehensive discussion to address these challenges and promising directions for future research.

\subsection{Future Research on Data Privacy}
\label{subsec:Data_Format}
\subsubsection{CT Medical Images}
In medical image analysis, there exists one work that implements the GAN-based model on MRI images to generate privacy-preserving synthetic medical images while maintaining segmentation performance.
However, in reality, CT images are more widely used in medical analysis than MRI images.
To leverage the advantages of GAN to further benefit medical image analysis in real applications, protecting private information in CT images with a performance guarantee would deserve researchers' attention.

\subsubsection{Sequential Records}
Although some methods have been proposed to protect public records' privacy before collaborative use, they take these records as discrete data for privacy-preserving processing.
These records actually are a kind of sequential data as the relation between two words in a textual sentence is not ignorable. 
Thus, separating one sentence into words may cause performance loss for data generation and privacy protection. 
In other words, it is indispensable to take these relations into account when generating privacy-preserving public records, which is one promising research direction for the effectiveness improvement in practice.

\subsubsection{Spatio-Temporal Information in Videos and Speeches}
In the prior works on privacy protection in videos and speeches, the sensitive information is hidden by noise added via a generator and the data utility is maintained by a discriminator through an adversarial training.
A video is treated as a sequence of image frames for generation, ignoring the spatio-temporal relation between frames; a speech is only considered as a distribution of voice information, overlooking the spatio-temporal relation between voice segments.
Notably, such spatio-temporal relations can be exploited as side-channel information to mine individuals' private information, especially with the development of deep learning models.
As a result, the challenge of incorporating spatio-temporal relations into privacy protection for videos and speeches should be overcome.

\subsubsection{Understanding Sensitive Information}
\label{subsec:Understanding}
Typically, the current GAN-based models add noise into the synthesized data to hide sensitive information through the generator and demonstrate the capability of privacy protection through the experimental results of reduced prediction/classification accuracy.
But, in these models, it is still unknown what type of sensitive information is hidden and where the noise is added.
This is because the generator is a black-box function of data synthesis, only relying on the discriminator's feedback in the adversarial training process.
Such a kind of training mechanism makes the privacy protection inefficient when facing the privacy detectors that are not taken into account by the discriminator(s) in the training process.
Therefore, understanding the privacy-related features in source data will be helpful to strengthen privacy protection in GAN.

\subsubsection{Guarantee of Privacy Protection}
\label{subsec:No_Guarantee_Privacy}
When applied to privacy-preserving data generation, all existing GAN-based methods fail to provide a theoretical guarantee of privacy protection.
The root cause is that the generator is a black-box function of data synthesis and its synthesis performance is mainly determined by the feedbacks of the discriminator during the adversarial training process.
What's worse, the adversarial training is not stable, resulting in the generator's unpredictable capabilities and the discriminator.
To further promote GAN' development and application, technique breakthrough is in desired to offer a theoretical guarantee of privacy protection.

\subsubsection{Computation Cost}
\label{subsec:Computation}
With the increased popularity of emerging applications, such as the Internet of Things (IoT) and edge/fog computing, a large amount of user data is shared through connected devices ({\em e.g.}, mobile phones), resulting in serious privacy leakage during transmission.
In order to protect data privacy in the connected devices timely before transmission, light-weighted GAN models are expected such that the computation cost ({\em e.g.}, time and energy) is affordable for these connected devices, in which balancing the tradeoff between computation performance and computation cost is an unavoidable challenge.

\subsection{Future Research on Model Privacy}

\subsubsection{GAN Model Improvement}
Future works on model privacy are tightly related to the GAN model's improvement, which is a mainstream research direction in the learning field and will continue to attract more research interest.
For example, the investigation on the convergence rate and mode collapse of GAN will definitely enhance GAN's efficacy on both attack and defense aspects.
Increasing the convergence rate would save more time and cost for attacker as well as provide a more efficient way to train a defense mechanism.
Rectifying the problem of mode collapse in GAN can yield a stronger generator with more representation capability, which can either encourage the attacker to recover data with higher quality when stealing preimage privacy or produce a more powerful denoising module for better defense.
To understand how to accelerate convergence and how to avoid mode collapse, conducting fundamental theoretical research is essential.

\subsubsection{GAN in Privacy Acquisition}
\label{sec:GAN_privacy_acquisition}
Using GAN to proceed privacy-related attack always faces a realistic problem: it requires a lot of real data as prior knowledge for training a well-performed GAN model, which is a strong assumption and is hard to be satisfied in practice.
Possible solutions to this problem include {\em transfer learning} and {\em probably approximately correct learning}.

Transfer Learning:
In real world, some private information is not accessible for the public, but lots of available related data can be used to bridge to privacy.
Transfer learning is a promising paradigm to perform knowledge transfer between public data and private data.
In the light of this idea, transfer learning and GAN can be integrated to establish an approximate GAN model that is very close to the GAN model trained on unknown private data.

Probably Approximately Correct Learning:
Until now, GAN has been applied to various attack methods, but there is no theoretical analysis illustrating why these methods can work and/or how good they can be.
It is probably approximately correct learning to provide a novel way of carrying out theoretical analysis on the above problems, filling the blank in literature.
As an elegant framework, probably approximately correct learning explores the mathematical analysis of machine learning and computational learning theory.
Especially, it can quantitatively analyze some parameters in learning algorithms, including the approximate correctness, the probability of getting approximate correctness, the number of sample needed in learning, and so on.
Since no work has studied probably approximately correct learning in GAN, there are many open questions for further investigation.
One novel idea is using series theory to derive the necessary conditions/requirement for launching privacy-related attack with GAN, which could tell us if it is worth launching an attack or not.
For example, what is the minimum size of training dataset to train a good GAN model with a certain attack success probability, and when there is limited training dataset locally, what is the upper bound of the attack accuracy.

\subsubsection{GAN in Privacy Protection}
Originally, GAN is born with a private feature: model separation to protect privacy.
That is, during the training process of GAN, only the discriminator can access data directly while the generator is innocent with data.
However, with the development of different attack methods, the privacy threats to GAN have been increased.

Differential Privacy:
Differential privacy has been treated as a golden defensive mechanism but has its defects, such as lower data utility when noise is accumulated.
In future work, investigating differential privacy deeply for GAN is still an attractive topic, for which some possible directions are briefly addressed in the following.
Firstly, according to the application requirements, different kinds of differential privacy could be explored, rather than only using $(\epsilon, \delta)$-differential privacy.
For example, Renyi-differential privacy and concentrated differential privacy are possible choices.
In~\cite{beaulieu2019privacy}, Renyi-differential privacy was proposed to achieve a more tighter bound on privacy loss.
As pointed by Aleksei and Boi~\cite{triastcyn2019federated}, bayesian differential privacy in federated learning is flexible and can save privacy loss significantly for utility critical applications.
Secondly, the convergence analysis of GAN in a differentially private setting should be studied. 
Current works mainly focus on the final results of trained models and rarely consider the convergence issues, leading GAN's research on an experiment-driven path.
In the long term, GAN's development needs the guidance of fundamental theory for accelerated progress in privacy protection.
Last but not least, the use of differential privacy should be granulated at an appropriate level for fine-grained protection, such as for instance level and client level.
It has been shown in previous work~\cite{hitaj2017deep} instance level differential privacy fails to protect preimage privacy.
Thus, in the design of differentially private GAN, more details should be considered for performance improvement.

Federated Learning:
Federated learning~\cite{mcmahan2017learning} can help relieve privacy leakage when facing a powerful attack.
Since the training data in federated learning is geographically distributed among all local non-contact clients, single failure issues can be avoided.
Also, integrating federated learning and differential privacy is not a hard job, which can further improve GAN's capability of privacy protection.

\subsection{Future Research on Security in GAN}

\subsubsection{Adversarial Sample Generation}
The adversarial samples generated by GAN can be used to either launch an attack or implement a defense, which has been intensively studied so far.

Attack:
It is worth mentioning that the actual attack success rate and training cost of GAN-based attack methods are not as good as those of the traditional optimization-based attack methods.
Improving these two metrics is closely related to GAN's fundamental research direction, {\em i.e.,} convergence and mode collapse.
The faster convergence rate, the less training cost; the lower mode collapse probability, the higher attack success rate.
Another reason for the low attack success rate is the lack of dedicated objective functions.
Among the generated results of a trained generator, some samples are benign while some are adversarial.
This is because the space of generative samples is a multi-fold space, where only a partial region may have adversarial attributes.
That is, even though the recall of adversarial samples is high, the precision of being adversarial in the entire space is quite small.
Thus, analyzing the generative space to restrict the probability of adversarial samples falling into the benign region from a mathematic perspective is a worth-thinking problem, which is non-trivial and requires thorough investigation.
In addition, when the number of training samples is limited, the generated adversarial examples become worse in the presence of a pure black-box classifier because black-box provides less useful information for training GAN.
To tackle this problem, as we illustrated in section~\ref{sec:GAN_privacy_acquisition}, probably approximately correct learning is a promising scheme to measure how much data is needed and/or how good the attack can be.

Defense:
Among the techniques of adversarial training, detection and de-noising, adversarial sample detection might be the most practical approach as it can be deployed in a plug-and-play mode without involving the trained models.
Motivated by this observation, GAN can be used to detect adversarial samples in various ways.
For example, we can configure $G$ as a feature squeezer that transforms data point $x$ to latent space $z$, {\em i.e.}, $G(x)=z$, and then train the discriminator $D$ as a binary classifier to check whether the latent vector comes from real data $G(x)$ or adversarial sample $G(x')$.
Finally, an adversarial sample could be easily detected through $D(G(\cdot))$ operation.

\subsubsection{Malware Related Research}

Most of the current malware related research based on GAN focuses on Android malware because the security protection mechanism is relatively consummate in the OS like Windows, Linux, and MacOS. 
This indicates that in the area of those OS, there still exist many research opportunities to overcome the unsolved challenges, such as analysis combining statistic and dynamic parameters, defensive programming, white-box attack, and code reviewing.

Statistic Analysis:
The majority of the previous works are about statistic analysis that effectively identifies the existing malware. 
With the help of GAN, the issue of insufficient malware samples can be fixed.
Besides, another drawback of statistic analysis on malware is its relatively weak performance of detecting unknowing malware. 
GAN is able to generate unknowing malware based on the known attacks but still needs experts to extract features by hand. 
How to detect malware against virtualization and extract statistic data more effectively make the issues worth exploring.

Dynamic Analysis:
Dynamic analysis is more robust than statistic analysis, but the existing dynamic analysis tools and techniques are imperfect. 
GAN can help enhance the performance of malware detection by applying dynamic analysis via generating more adversarial samples. 
But, only several dynamic features are actually utilized in the current works. 
As virtualization is getting widespread, dynamic analysis on VMM and hypervisor may be a 
promising direction.
Moreover, dynamic analysis on the side channel of attack behaviors is still left blank now, which should be filled to advance malware detection techniques.

White-box Attacks:
Black-box attacks and semi white-box attacks are the focus of most of the existing works.
Compared with black-box attacks, white-box attacks require higher transparency of the target systems. 
Usually, white-box attacks are related to code review, which has a high requirement for experts' experience.
Besides, there is a lack of datasets for the white-box attack, which is a good application scenario for GAN.

\subsubsection{Bioinformatic Recognition}

Bioinformatic based recognition has been widely used in various areas, some of which are related to the applications requiring a high degree of security protection, such as bioinformatic authentication. 
Slight modifications on bioinformatic may cause the results of recognition change thoroughly. 
Current works are limited to fingerprint and face recognition, ignoring other important bioinformatic like iris.
As a result, more efforts are needed to design GAN-based methods for different bioinformatic recognition systems.

\subsubsection{Industrial Security \& Others}

Industrial security is more complex and challenging and thus has higher requirements. 
Here, using GAN to generate adversarial data from limited data is helpful in enhancing industrial security.
However, GAN also has its limitation: the probability distribution of adversarial data may be unbalanced when there is too little original data. 
Especially in a situation where the specific context of a test is missing, combining transfer learning and GAN may enable the generation of adversarial samples from limited data resources.

\section{Conclusion}
\label{sec:conclusion}
This survey intensively reviews the state-of-the-art approaches using GAN for privacy and security in a broad spectrum of applications, including image generation, video event detection, records publishing, distributed learning, malware detection, fraud detection, and so on.
For the different purposes of attack and defense, these existing approaches establish problem formulation based on the variants of GAN framework, taking into account attack success rate, classification/prediction accuracy, data utility, and other performance metrics.
After a thorough analysis, the unsolved challenges and promising research directions are provided for further discussion from perspectives of application scenario, model design, and data utilization.

\begin{acks}
	This work is partially supported by the National Science Foundation under Grant (No.1741277, No.1829674, No.1704287, No.1912753 and No.2011845).
\end{acks}

\bibliographystyle{ACM-Reference-Format}
\bibliography{GAN_survey}

\end{document}